\documentclass{article}
\usepackage[nonatbib,preprint]{neurips_2022}

\usepackage[utf8]{inputenc} 
\usepackage[T1]{fontenc}    
\usepackage{hyperref}       
\usepackage{url}            
\usepackage{booktabs}       
\usepackage{amsfonts}       
\usepackage{nicefrac}       
\usepackage{microtype}      
\usepackage{xcolor}         

\usepackage{placeins}
\usepackage{algorithm}
\usepackage{algorithmic}
\usepackage{array}
\usepackage{overpic}
\usepackage{wrapfig}
\usepackage{bm}
\usepackage{comment}
\usepackage{pgfplots}
\usepackage{pgf}
\usepackage{tikz}
\usepackage{soul}
\usepackage{tabstackengine}
\usepackage{mathtools}
\usepackage{multirow}
\usepackage{caption}
\usepackage{scalerel}[2014/03/10]
\usepackage{stackengine,wasysym}

\usepackage{amssymb}
\usepackage{amsmath}
\usepackage{amsthm}
\newtheorem{thm}{Theorem}[section]
\newtheorem{lemma}[thm]{Lemma}

\newcommand{\M}{\mathcal{M}}
\newcommand{\N}{\mathcal{N}}

\newcommand{\EmbNet}{\mathcal{E}}
\newcommand{\FeatNet}{\mathcal{F}}
\newcommand{\OurBasis}{{\Phi^{\EmbNet}}}
\newcommand{\LBOBasis}{{\Phi^{\Delta}}}

\newcommand{\DiffNet}{GFM_{DN}}
\newcommand{\Snob}{\widetilde{S}}
\newcommand{\Fnob}{\widetilde{F}}
\newcommand{\Ls}{L_{E_D}}
\newcommand{\Lo}{L_{\bot}}
\newcommand{\Low}{L_{w\bot}}
\newcommand\dwidetilde[1]{\ThisStyle{%
  \setbox0=\hbox{$\SavedStyle#1$}%
  \stackengine{-.1\LMpt}{$\SavedStyle#1\,\,$}{%
    \stackengine{\dimexpr-3.5\LMpt+.3pt}{%
    \stretchto{\scaleto{\mkern.2mu\AC}{.5150\wd0}}{.7\ht0}%
    }{%
    \stretchto{\scaleto{\mkern.2mu\AC}{.5150\wd0}}{.7\ht0}%
    }{O}{c}{F}{T}{S}
  }{O}{c}{F}{T}{S}%
\!\!}}

\def\test#1{$%
  \scriptstyle\dwidetilde{#1}\
$}

\DeclareMathOperator*{\argmin}{arg\,min}

\newcommand{\rick}[1]{\textcolor{black}{#1}}

\definecolor{mygray}{RGB}{120,120,120} 

\renewcommand{\arraystretch}{1.2}
\newcommand{\ra}[1]{\renewcommand{\arraystretch}{#1}}

\title{Smoothness and effective regularizations in learned embeddings for shape matching}

\author{Riccardo Marin\\
Sapienza \\ University of Rome\\
{\tt\small marin@di.uniroma1.it}
\and
Souhaib Attaiki\\
LIX\\ Ecole Polytechnique, IP Paris\\
{\tt\small attaiki@lix.polytechnique.fr}
\and
Simone Melzi\\
Sapienza \\ University of Rome\\
{\tt\small melzi@di.uniroma1.it}
\and
Emanuele Rodol{\`a}\\
Sapienza \\ University of Rome\\
{\tt\small rodola@di.uniroma1.it}
\and
Maks Ovsjanikov\\
LIX\\ Ecole Polytechnique, IP Paris\\
{\tt\small maks@lix.polytechnique.fr}
}

\begin{document}

\maketitle

\begin{abstract}
Many innovative applications require establishing correspondences among 3D geometric objects. However, the countless possible deformations of smooth surfaces make shape matching a challenging task. Finding an embedding to represent the different shapes in high-dimensional space where the matching is easier to solve is a well-trodden path that has given many outstanding solutions.
Recently, a new trend has shown advantages in learning such representations. This novel idea motivated us to investigate which properties differentiate these data-driven embeddings and which ones promote state-of-the-art results. In this study, we analyze, for the first time, properties that arise in data-driven learned embedding and their relation to the shape-matching task. Our discoveries highlight the close link between matching and smoothness, which naturally emerge from training. Also, we demonstrate the relation between the orthogonality of the embedding and the bijectivity of the correspondence. Our experiments show exciting results, overcoming well-established alternatives and shedding a different light on relevant contexts and properties for learned embeddings.
\end{abstract}


\section{Introduction}
\label{sec:introduction}
Deep learning largely bases its success on converting input features into others useful to solve specific applications. From a more abstract perspective, training a neural network is no more than optimizing the parameters of a non-linear function which maps between two embeddings. Scholars largely explored this perspective, showing also how operations like interpolation \cite{marin2021spectral}, algebra \cite{santa2018neural}, or analogies \cite{yang2019deep} difficult in the starting domain, may be simple in learned onece. Inferring and manipulating relations between different objects lies at the core of learning: deciding what is different, what is similar, and what are the relations in a population eases the learning process and enables powerful applications. Several fields uses alignment in learned embedding as a way to infer relations; few examples among others are Natural Language Processing \cite{pado2009robust, shi-xiao-2019-modeling, wang-etal-2018-cross-lingual}, Biology \cite{zheng2019sense, nelson2019embed}, and 3D Geometry \cite{torgerson1952multidimensional,bronstein2006generalized,wu20153d}. This latter domain, given its nature, often paves the road with novel tools, providing geometrical insights that quickly spread across communities \cite{CPDNLP, ravindra2019rigid, bronstein2017geometric}.

In 3D shape analysis, point-wise features and descriptors as high-dimensional embeddings are analyzed for decades \cite{loncaric1998survey, AWFT, sun2009concise}. One practical approach, namely the multidimensional scaling~\cite{torgerson1952multidimensional}, consists in embedding the intrinsic geometry of a 3D shape into a higher-dimensional space where Euclidean distances approximate the proper metric of the object. The convenience of this procedure is clear, as one commonly prefers to work with extrinsic distances (lengths of straight lines in Euclidean space) instead of intrinsic ones (lengths of curved paths on a surface). Another possibility arises from the eigenfunctions of the Laplace-Beltrami Operator (LBO) that have played a prominent role in deformable shape representation. Their theoretical properties (i.e. orthogonality and smoothness) make them convenient for several applications, such as surface fairing, function analysis, segmentation, and shape matching \cite{field1988laplacian,reuter2009discrete,ovsjanikov2012functional}. Among the variations on this theme \cite{choukroun2016hamiltonian,LMH,CMH,melzi2019sparse}, recently, some learned embeddings specific for shape matching have been proposed \cite{wang2019deep, smirnov2021hodgenet}. Along this line, \cite{LIE2020} proposed to learn an alternative embedding to address non-rigid shape matching for point clouds. It offers a novel data-driven direction and provides promising results. Despite the wide attention devoted to these techniques, the research mainly follows general intuitions without analyzing how such properties may influence the relations between elements in the learned embedding.

We aim to fill this knowledge gap by analyzing some properties and their effectiveness in the learned embedding, which kind of relations they promote, and their applicative contexts.

\begin{figure*}
\centering
			\begin{overpic}[trim=0cm 0cm 0cm 0cm,clip, width=0.80\linewidth]{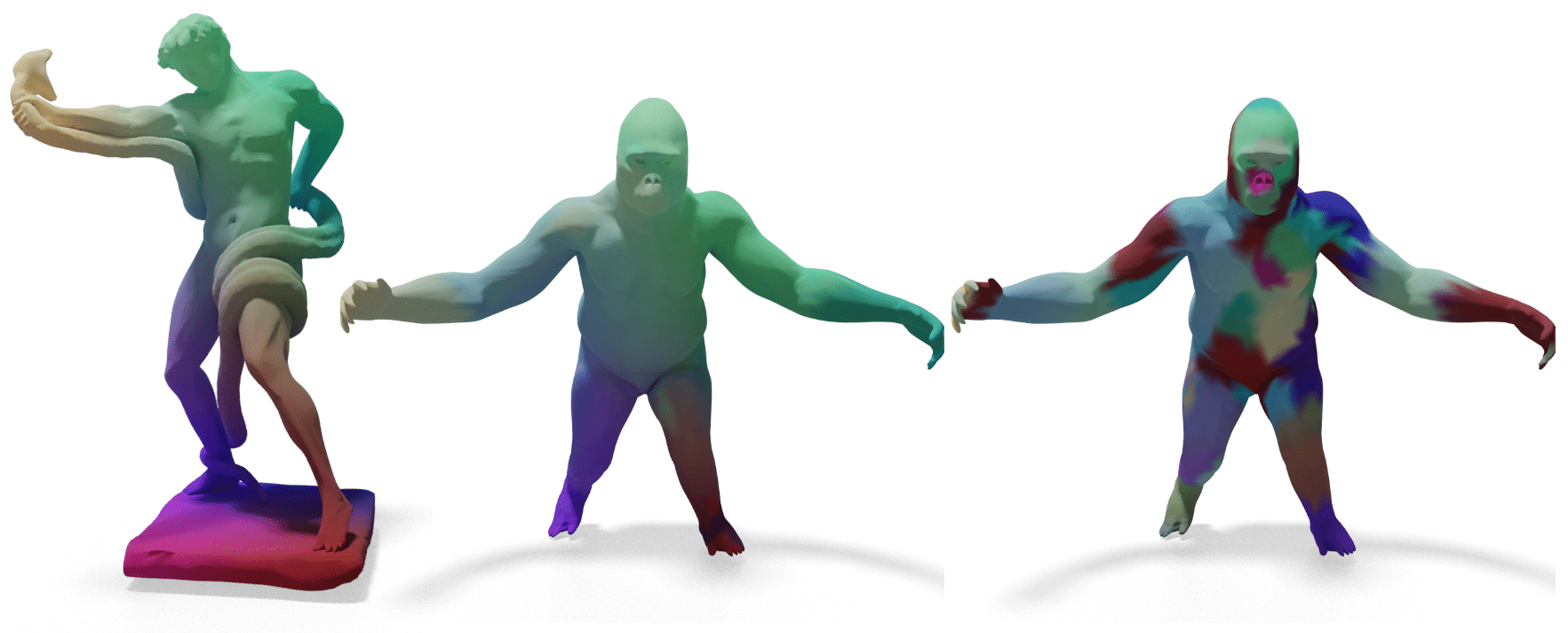}
			\put(15,0){$\M$}
			\put(40,0){Our}
			\put(75,0){$GFM_{DN}$}
			\end{overpic}
\caption{\label{fig:teaser} An example of extreme non-isometric matching. The left shape is a statue from \cite{scantheworld}, the second is a gorilla from \cite{TOSCA}.}
\end{figure*}

We instantiate the method from \cite{LIE2020} to work on meshes adopting a state-of-the-art feature extractor \cite{sharp2020diffusion}; this setting lets us define interesting properties clearly (e.g., smoothness). Moreover, the point cloud scenario of \cite{LIE2020} limits the possible analysis given the absence of reliable axiomatic competitors.
For the first time, we study the effect of smoothness and orthogonality on features for matching, providing both theoretical derivations and empirical evidence of our results. Furthermore, previous methods mainly focus on objects in bijection or with weak remeshing. Instead, we consider both bijective and non-bijective cases, showing more realistic scenarios. Our analysis not only pursues theoretical considerations but also has an applicative impact. Designing the embedding properties carefully improve performances, achieving a precision level otherwise impossible with previous representations (even when they enjoy the best transformation possible). Figure \ref{fig:teaser} is a qualitative example of this. 
We believe that our results significantly impact any field where embedding alignment is a common practice, starting from the Geometry Processing one where this problem is preminent.

\rick{To summarize, our work provides the following novelties and contributions:}
\begin{enumerate}
    \item We suggest \emph{novel effective regularizations} to learn embedding for the matching task. We introduce new simple losses, providing theoretical motivations and showing their practical utility. 
    \item We offer an \emph{extensive analysis of learned embeddings} in the shape matching application. We question some of the most established beliefs from the 3D shape matching community for the first time. We shed new light on prominent properties that characterize existing solutions disclosing new research direction.
    \item We \emph{outperform state-of-the-art alternatives}. We exceed strong shape-matching competitors in multiple scenarios. To our best knowledge, this is the first study that produces an embedding that can represent shape correspondence better than LBO eigenfunctions, even for the near-isometric case.  
    
\end{enumerate}

All the data and code will be made publicly available for research purposes.

\section{Related work}
\label{sec:related}

Shape matching has a long history \cite{van2011survey,sahilliouglu2020recent}. We cover the more closely related methods to our work.

\paragraph{Embeddings for shape matching} Multidimensional scaling and its variants \cite{torgerson1952multidimensional,bronstein2006generalized,bronstein2006multigrid} are seminal approaches for computing shape embeddings for shape matching. The main idea is to realize the geodesic metric of a given $3$D object as the Euclidean (also called restricted) metric of some high-dimensional space. Such approaches are variational since, in general, a solution does not exist even for simple cases \cite{linial2003finite}. Several works have recently proposed extensions of the Laplacian eigenbasis to deal with difficult settings. In \cite{LMH}, it was proposed to augment the LBO set with a locally supported basis, defined only in a specific region, to increase its local representation capability; \cite{CMH} proposes to add extrinsic information by appending the orthonormalized version of the three coordinates vectors; \cite{kovnatsky2013coupled} suggested finding a basis by jointly diagonalizing the Laplacians of the input pair of shapes. Interestingly, all the previous methods base their representation on the standard LBO basis.

Another popular approach in shape matching is finding a set of features (also called descriptors) that are able to identify each point uniquely, such as the Global Point Signature \cite{rustamov2007laplace}, Wave Kernel Signature \cite{aubry2011wave}, Heat Kernel Signature \cite{sun2009concise}, AWFT \cite{AWFT}, SHOT \cite{SHOT} and its robust version proposed in \cite{GFRAMES}. However, while they are widely used, such features alone are insufficient to provide accurate point-to-point matching, especially in non-rigid scenarios.

\paragraph{Functional maps} \cite{ovsjanikov2012functional,ovsjanikov2017computing} introduce and describe the paradigm considered in this paper. They move the correspondence problem to the functional domain. The advantages come from the low-dimensional approximation of these spaces, which arises from a set of fixed basis functions (mainly the LBO eigenvectors). This framework has been extended to take into account partiality \cite{rodola2017partial}, to provide precise matching within triangle faces \cite{ezuz2017deblurring}, and improved with several regularizations ~\cite{nogneng2017informative,ren2018continuous} or iterative refinement~\cite{melzi2019zoomout,MapTree,pai2021fast}. 

\paragraph{Learning-based pipelines} Several data-driven pipelines for shape matching only address the rigid case \cite{pais20203dregnet,Huang_2020_CVPR,sarode2019pcrnet, wang2019deep}, while 3DCoded \cite{groueix20183d} addresses the non-rigid setting at the cost of an expensive optimization at test time. In the latter setting, an entire line of research has been devoted to learning the descriptors used by functional maps. This has been done using random forests~\cite{rand}, and more recently via deep learning models such as FMNet~\cite{litany2017deep}. Recently, \cite{donati2020deep} proposed to learn the features directly from the point cloud coordinates while keeping the LBO basis. The main inspiration of our work is~\cite{LIE2020}, which proposes to learn both the bases and descriptors. This method only processes 3D point clouds since it uses the PointNet architecture \cite{qi2017pointnet} without exploiting the possible input connectivity. Furthermore, it does not investigate the properties and applicability of the learned basis. Among the other contributions, our work also aims to fill this gap.
\section{Notation and background}
\label{sec:notation}
\paragraph{3D shapes.} We model 3D shapes as 2-dimensional Riemannian manifolds embedded in $\mathbb{R}^3$. In the discrete setting, we encode these objects as triangular meshes composed of vertices and faces defined by the oriented triplet of vertices that belong to the same triangular face. Each face is glued to a maximum of three other faces, one for each edge.
We denote with $\M$ and $\N$ a pair of shapes and with $X_{\M}\in \mathbb{R}^{n_{\M}\times 3}$ and $X_{\N}\in \mathbb{R}^{n_{\N}\times 3}$ the list of the 3D coordinates of their $n_{\M}$ and $n_{\N}\in \mathbb{N}$ vertices. 
To each of these shapes, we associated the Laplace-Beltrami Operator (LBO), the second-order partial differential operator extending the standard Laplacian to non-Euclidean domains and denoted respectively as $\Delta_{\M}$ and $\Delta_{\N}$. We adopt the same notation referring to the square matrices, with size $n_{\M}$ and $n_{\N}$ respectively, which encode these operators in the discrete setting and that can be estimated through the cotangent weight formula \cite{pinkall1993computing,meyer03}.

The LBO is a symmetric, positive semi-definite operator which admits an eigendecomposition with non-negative real eigenvalues, sorted in non-descending order $0=\lambda_1\leq \lambda_2 \leq \ldots $. The eigenfunctions $\phi_1^{\Delta}, \phi_2^{\Delta}, \ldots $ associated with each eigenvalue compose a basis for the space of square-integrable functions defined over the surface, in analogy with the Fourier basis on Euclidean domains.
In the discrete setting, each eigenfunction corresponds to a vector with a length equal to the number of vertices.
We store the set of the $k$ eigenfunctions associated to the first $k$ eigenvalues with smallest absolute values, as columns of a matrix $\LBOBasis = [\phi_1^{\Delta}, \ldots , \phi_k^{\Delta}]$. Each \textit{row} of this matrix is a vector in $\mathbb{R}^{k}$ and is referred to as the \textit{spectral embedding} of the corresponding vertex of the mesh. The matrix $\LBOBasis$ thus encodes the spectral embedding of the entire shape.

%
\paragraph{Functional maps.} Given the two shapes $\M$ and $\N$, together with their truncated set of LBO eigenfunctions $\LBOBasis_{\M}$ and $\LBOBasis_{\N}$ respectively, we denote with $\mathcal{F}(\M)$ and $\mathcal{F}(\N)$ the space of real-valued functions defined on $\M$ and $\N$.
Any point-to-point correspondence $T_{\M\N}:\M\rightarrow\N$ induces a functional mapping (with opposite direction) $T^{\mathcal{F}}_{\N\M}:\mathcal{F}(\N)\rightarrow\mathcal{F}(\M)$ via pull-back.  
Exploiting the Fourier analogy, we can approximate the spaces $\mathcal{F}(\M)$ and $\mathcal{F}(\N)$ in the given bases $\LBOBasis_{\M}$ and $\LBOBasis_{\N}$ of size $k$.
Thanks to this approximation, we can compactly encode the mapping $T^{\mathcal{F}}_{\N\M}$ in a matrix $C_{\N\M} \in \mathbb{R}^{k\times k}$ which corresponds to the linear transformation that maps the coefficients of functions approximated by $\LBOBasis_{\N}$ to the coefficients of their images through $T^{\mathcal{F}}_{\N\M}$ represented by $\LBOBasis_{\M}$. In matrix notation, if we encode the point-to-point map $T_{\M\N}$ in a binary matrix $\Pi_{\M\N} \in \mathbb{R}^{n_{\M} \times n_{\N}}$, such that its entries $\Pi_{\M\N}(i,j) = 1 $ if and only if the correspondence $T_{\M\N}$ associates, to the $i$-th vertex of $\M$, the vertex of index $j$ on $\N$, then we can explicitly compute the functional map $C_{\N\M} = \LBOBasis_{\M}^{\dagger}\Pi_{\M\N}\LBOBasis_{\N}$, where we denote with $\dagger$ the Moore–Penrose pseudoinverse. 
In this framework, due to the analogy with Fourier, the matrices $\LBOBasis_{\M}$ and $\LBOBasis_{\N}$ play the role of bases for the functional spaces. For this reason, we will refer to them both as spectral embedding and as basis.

\paragraph{Linearly invariant training.}
Our work follows the architecture introduced in~\cite{LIE2020} but modifies it accordingly to work with triangular meshes.
Given a set of shape pairs $(\M,\N)$, equipped with a ground truth correspondence $\Pi^{gt}_{\M\N}$, we train an embedding network $\EmbNet$. This network takes as input the coordinates $X_{\M}$ of a shape $\M$ and outputs a high-dimensional embedding $\OurBasis_{\M}$.
The network is trained by considering the optimal linear transformation between the two shape embeddings: 
\begin{equation}
\label{eq:defcopt}
    C^{opt}_{\N\M} = (\OurBasis_{\M})^{\dagger} \Pi^{gt}_{\M\N}\OurBasis_{\N}\,,
\end{equation}
which is converted to a penalty measuring how well the embedded points are aligned; since nearest-neighbor is not differentiable, we cast this problem as follows:
\begin{align}
    D &= dist( \OurBasis_{\M}  C^{opt}_{\N\M}, \OurBasis_{\N}) \\
    S_{\M\N} &= softmax(-D)\,,
\end{align}
where $dist$ computes the matrix of Euclidean distances in the embedding space, and $S_{\M\N}$ acts as a score of similarity between points. Finally, the network loss is formulated as:
\begin{equation}
    L_{\EmbNet}(\OurBasis_{\M}, \OurBasis_{\N}) = \sum \|S_{\M\N} X_\N - \Pi^{gt}_{\M\N} X_\N \|_2^2\,.
    \label{eq:loss}
\end{equation}
In experiments which also require descriptors, we learn them following \cite{LIE2020} paradigm: we train the same architecture of $\EmbNet$, with the loss proposed by Equation 4 of \cite{LIE2020}.

\rick{\paragraph{DiffusionNet.} We adopt a state-of-the-art feature extractor as the backbone. DiffusionNet \cite{sharp2020diffusion} relies on a diffusion process over the surface to propagate the features information. A diffusion block is based on learning the parameters of a heat diffusion process, plus an anisotropic filter. We refer to the related paper for the technical details. 
We choose DiffusionNet since it is one of the most recent and promising architectures designed to exploit the structure of the $3$D surfaces.}

\paragraph{Notataion.} We denote with $\OurBasis$ the embedding learned by minimizing $L_{\EmbNet}$ and with  $\LBOBasis$ the LBO basis (both coincide with a matrix), while we adopt \emph{Our} when we use both learned embedding and descriptors. 

\begin{figure*}[!t]
	\vspace{0.5cm}
	
	\footnotesize

    \begin{tabular}{c}
    \hspace{0.7cm}	
	\begin{minipage}{\linewidth}

	\begin{overpic}[trim=0cm 0cm 0cm 0cm,clip, width=0.92\linewidth]{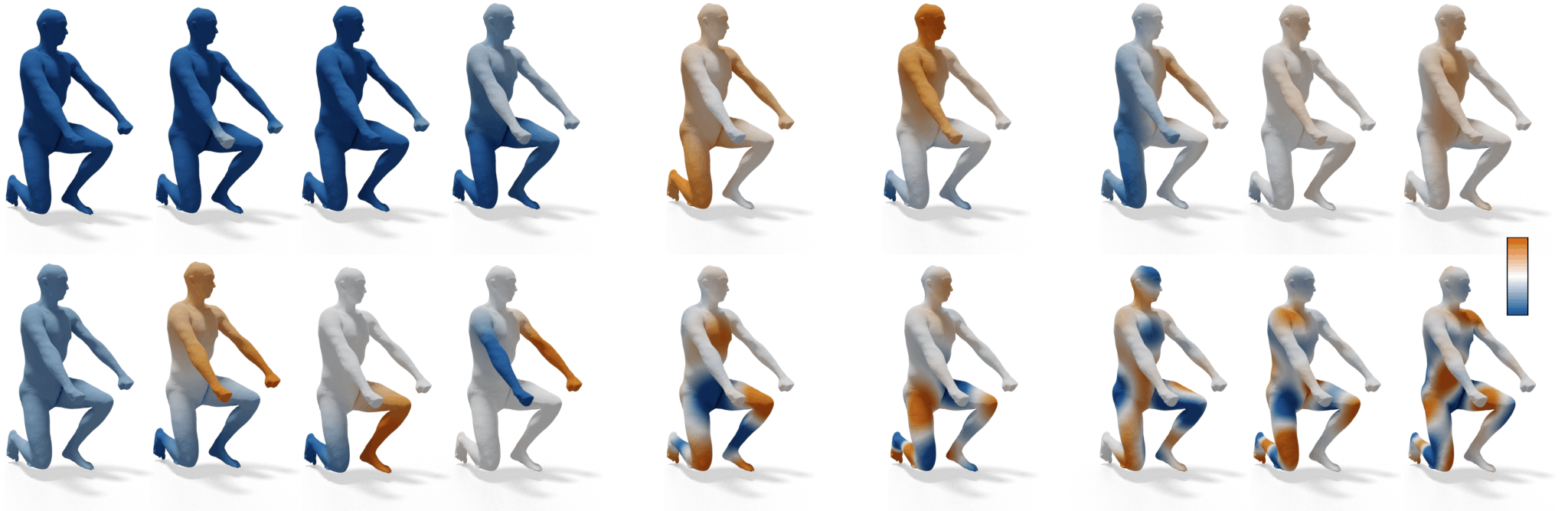}
		\put(-3,22){\scriptsize $\OurBasis$}
		\put(-3,10){\scriptsize $\LBOBasis$}
		\put(-7.6,30.6){\scriptsize $E_{D}(\OurBasis)$}
		\put(2,33.5){\scriptsize{$0.62$}}
		\put(12,33.5){\scriptsize{$1.18$}}
		\put(20,33.5){\scriptsize{ $1.94$}}
		\put(30,33.5){\scriptsize{$3.02$}}
		
		\put(43,33.5){\scriptsize{$9.63$}}
		\put(56,33.5){\scriptsize{$19.27$}}
		
		\put(71,33.5){\scriptsize{$90.97$}}
		\put(81,33.5){\scriptsize{$145.51$}}
		\put(91,33.5){\scriptsize{$160.99$} }

		\put(39,10){{\dots} }  
		\put(38,25){{\dots} }  
		\put(53,10){{\dots} }  
		\put(52,25){{\dots} }  
		\put(65,10){{\dots} }  
		\put(65,25){{\dots} }  

        \put(98,11.95){{\tiny{$-\sigma$}} } 
        \put(98,14.4){\tiny{$0$} }
        \put(98,16.95){\tiny{$+\sigma$} } 
        
        \put(-7.6,0.3){\scriptsize{$E_{D}(\LBOBasis)$}}
		\put(3,-0.5){\scriptsize{$0$}}
		\put(12,-0.5){\scriptsize{$2.58$}}
		\put(20,-0.5){\scriptsize{$3.44$}}
		\put(30,-0.5){\scriptsize{$5.04$}}
		
		\put(43,-0.5){\scriptsize{$87.84$}}
		
		\put(56,-0.5){\scriptsize{$156.95$}}
		
		\put(71,-0.5){\scriptsize{$221.27$}}
		\put(81,-0.5){\scriptsize{$226.18$}}
		\put(91,-0.5){\scriptsize{$239.27$}}
		
		\put(3,-4.1){\scriptsize{$\mathbf{1^{th}}$}}
		\put(12,-4.1){\scriptsize{$\mathbf{2^{nd}}$}}
		\put(22,-4.1){\scriptsize{$\mathbf{3^{rd}}$}}
		\put(32,-4.1){\scriptsize{$\mathbf{4^{th}}$}}
		\put(45,-4.1){\scriptsize{$\mathbf{18^{th}}$}}
		\put(58,-4.1){\scriptsize{$\mathbf{28^{th}}$}}
		\put(73,-4.1){\scriptsize{$\mathbf{38^{th}}$}}
		\put(83,-4.1){\scriptsize{$\mathbf{39^{th}}$}}
		\put(93,-4.1){\scriptsize{$\mathbf{40^{th}}$}}
	\end{overpic}
	\vspace{0.4cm}
	
	\end{minipage}
    \end{tabular}
	\caption{We visualize the dimensions of the embeddings as a function over the surface. For each one, we report its Dirichlet energy $E_D$ as a measure of the smoothness.
	}
	\label{fig:basis}	
\end{figure*}
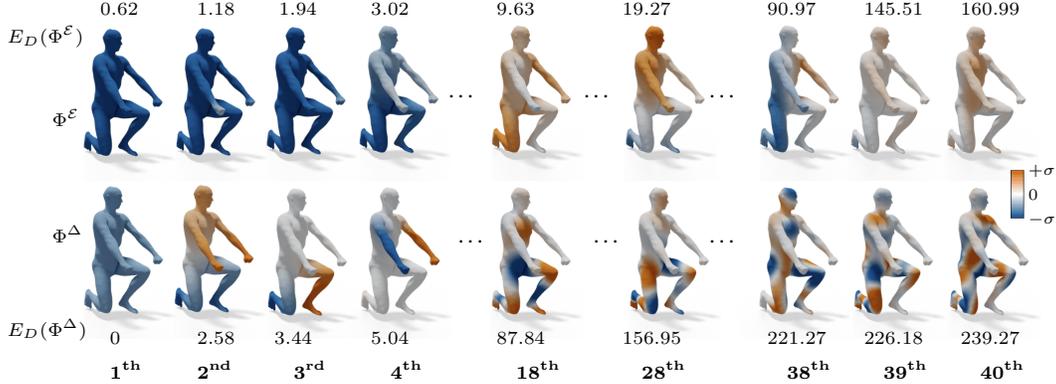

\section{Method}
\label{sec:method}

The main purpose of~\cite{LIE2020} was proposing a pipeline end-to-end (learning basis and descriptors) for sparse point clouds matching, where no good basis alternatives are known and require thousands of training data. We emphasize that our study is orthogonal to ~\cite{LIE2020} and other approaches. 
Our work does not aim to introduce a novel framework but to analyze properties of existing ones by exploring different basis in a data-driven implementation similar to ~\cite{LIE2020}. Remarkably, we propose a novel perspective, a challenging setting, and a far more ambitious goal: overtaking LBO in its domain, with a limited training set, studying which properties make it competitive and which can be relaxed. We focus on the bases, considering descriptors only in application experiments.
This approach lets us unveil theoretical properties and insights about good representations for embedding alignment.

The main properties of $\LBOBasis$ are \emph{orthogonality} and \emph{smoothness}.  
Furthermore, these functions carry a natural order given by the corresponding eigenvalues.
We describe the losses we adopt to force the learned basis to own the desired properties. All these losses do not require supervision at training time. 

\paragraph{Smoothness.}
A good point-wise representation for the matching application should provide similar encoding for the nearest points. Smoothness could guarantee this property as preliminary proved in ~\cite{LIE2020}.
In the functional context, smoothness coincides with the Dirichlet energy (namely $E_{D}$): the smaller the energy, the smoother the function. We refer to the supplementary materials for its discretization.
We foster this property through the following loss:
\begin{equation}
\Ls(\OurBasis_{\M}) = \sum \| diag(\OurBasis_{\M}^T A \LBOBasis \Lambda \LBOBasis^T A \OurBasis_{\M})\|_2^2
\label{eq:smooth}
\end{equation}
$\Ls$ is a dense version of the Dirichlet energy, and thus its minimization promotes smoothness. We will show in our analysis the prominent role of smoothness in functional shape matching. A visualization of functions with different smoothness levels is depicted in Figure \ref{fig:basis}.

\paragraph{Orthogonality.}
Given a multi-dimensional representation of the points, looking at them a vertex-wise functions (i.e. columns of the embedding matrix), we impose orthogonality by minimizing the following loss:
\begin{align}
\Lo(\OurBasis_{\M})  & = \sum \| \OurBasis_{\M}^{\top}\OurBasis_{\M} - Id\|_2^2 , \label{eq:ortho}
\end{align}
where $Id$ is the identity matrix.
In the Appendix \ref{sec:appendixA}, we prove that the orthogonality of the learned features promotes \emph{weak bijectivity} in the matching. 
We refer to $\Low$ when the loss is scaled by a $10^{-10}$ factor.

\paragraph{Sparsity.}  
We do not limit our analysis to the properties owned by the LBO basis. Inspired by the great success of the \emph{compressed sensing} in signal processing, we introduce a new loss to promote sparsity in the learned embedding. Given the importance of smoothness for the matching, we only combine the sparsity and the smoothness losses as follows: 
\begin{align}
L_{\ell1+E_D} = L_{\ell1} + w_{E_D} \Ls, \\
\mbox{where } L_{\ell 1}(\OurBasis_{\M}) = \sum \|\OurBasis_{\M}\|_1.
\label{eq:l1s}
\end{align}
Due to the different magnitude of the losses, in $L_{\ell1+S}$ we scaled $Loss_{S}$ by $w_{E_D}=10^6$. We highlight that, to assess the role of the matching loss in the proposed procedure, we do not optimize for the loss in Equation \ref{eq:loss} when we adopt the $L_{\ell1+E_D}$ one.

\section{Dataset and settings.}
\label{sec:settings}
This section introduces the benchmarks and the parameters we select in our experiments.

\paragraph{Bijective.} We namely refer to \emph{bijective} setting when the shapes in the train and in the test share the same mesh and thus are in 1:1 correspondence. We consider two human datasets: SCAPE \cite{anguelov2005scape} (12500 vertices) and FAUST \cite{FAUST} (6890 vertices). As done in previous work \cite{sharp2020diffusion}, we the train on one dataset and test on the other. We denote with \textbf{F+S} when we train on FAUST \cite{anguelov2005scape} and test on SCAPE \cite{FAUST}. \textbf{S+F} refers to the opposite. From now on, we adopt \textbf{+} to indicate training (on the left) and test set (on the right). Furthermore, we consider the SMAL dataset \cite{SMAL}, composed of animal shapes with pose variations and from five species (big cats, canines, ovine, bovine, and hippos). The training set comprises 25 shapes, which is tight compared to the standard ones. The test set includes 300 pairs of 25 shapes unseen at training time. This setting, namely \textbf{SMAL}, includes strong non-isometric pairs.

\paragraph{Non-bijective.} In this scenario, we take pairs of shapes that do not share the same connectivity at training or inference time. We indicate with $\Snob$ when they come from SCAPE, one at a full resolution and one remeshed with the $25\%$ of the vertices. Similarly, we use $\Fnob$ for FAUST.

\paragraph{Different densities between train and test.} Finally, with \test{S+F}, we denote the case in which, at training time, both shapes from SCAPE have a quarter of vertices, but at test time, we used the full resolution shapes from $F$. In this case, the discretization density at training differs from one of the test shapes. We choose this setting to emphasize the non-bijectivity since $F$ is less isometric than $S$.

\rick{\paragraph{Settings.} If not differently stated, we train our embedding networks to output $40$-dimensional embeddings. If descriptors are used, they are $80$ functions obtained as described in Section \ref{sec:notation}. Since \cite{LIE2020} does not describe the relationship between basis and descriptors cardinalities, this choice is motivated by an analysis that we report in the Supplementary Material.} 
In our quantitative evaluation, we compute the average of the geodesic error measured for each point as the geodesic distance between the estimated correspondence and the ground truth one. To facilitate the comparison, we visualize all the shapes with a similar orientation, but several rigid transformations in the 3D space occur among the shapes both at training and test time. During training, we do data augmentation by applying rotations on $y$-axis. We refer to the Supplementary Material for details on datasets, evaluation, training, and testing procedures.

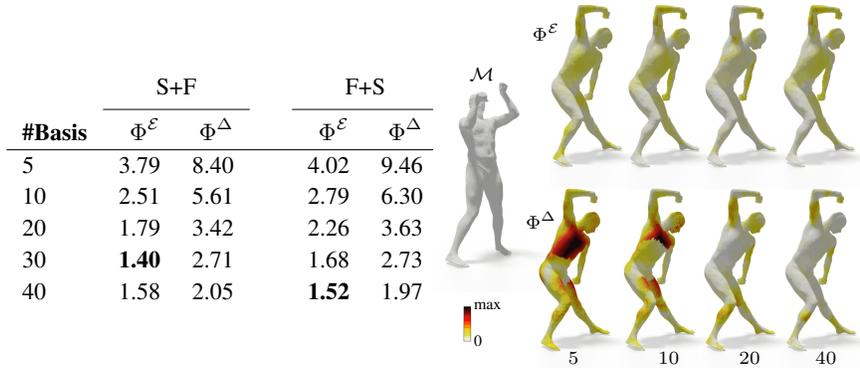
\begin{figure*}[!t]
	\centering
	\begin{tabular}{cc}
		
		\hspace{-0.3cm}
		\begin{minipage}{0.35\linewidth} 
            \footnotesize
        \begin{tabular}{lrrcrrr}
             & \multicolumn{2}{c}{S+F} & \phantom{a}  &\multicolumn{2}{c}{F+S} \\
             \cmidrule{2-3} \cmidrule{5-6} 
        \textbf{\#Basis} & $\OurBasis$            & $\LBOBasis$   & & $\OurBasis$             & $\LBOBasis$    \\
        \hline
        5        & 3.79            & 8.40  & & 4.02            & 9.46  \\
        10       & 2.51            & 5.61  & & 2.79            & 6.30  \\
        20       & 1.79            & 3.42  & & 2.26            & 3.63  \\
        30       & \textbf{1.40}   & 2.71  & & 1.68            & 2.73  \\
        40       & 1.58            & 2.05  & & \textbf{1.52}   & 1.97 
        \end{tabular} 
                    
		\end{minipage}
		& 
		\begin{minipage} {0.45\linewidth} 
			\hspace{0.3cm}
			\begin{overpic}[trim=0cm 0cm 0cm 0cm,clip, width=0.95\linewidth]{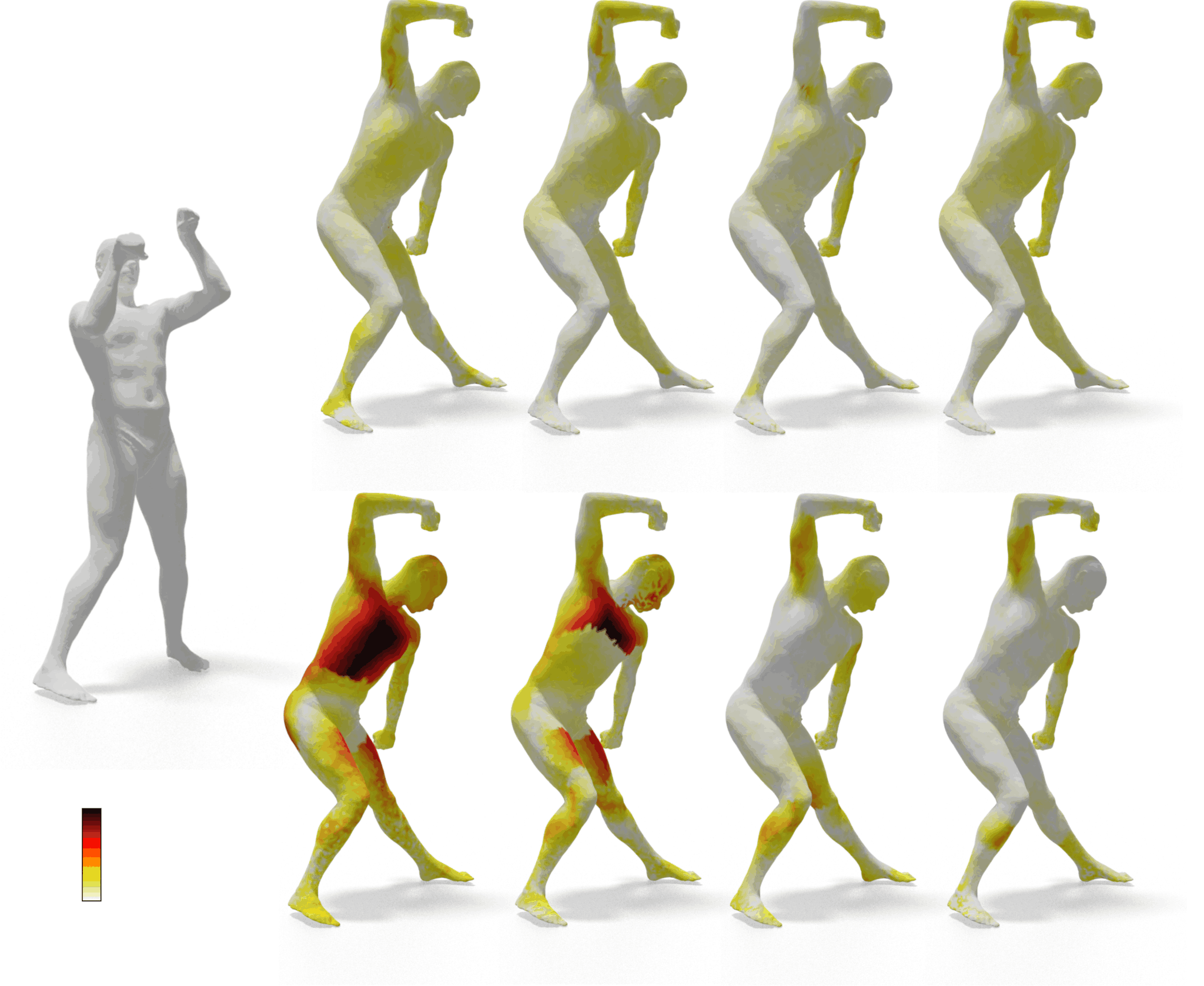}
				\put(8,65){\scriptsize{$\M$}}
				\put(22,74){\scriptsize{$\OurBasis$}}
				\put(20.5,32){\scriptsize{$\LBOBasis$}}
				
				\put(9,14){\tiny{max}}
				\put(9,6){\tiny{0}}
				
				\put(30,2){\scriptsize{$5$}}
				\put(50,2){\scriptsize{$10$}}
				\put(68,2){\scriptsize{$20$}}
				\put(85,2){\scriptsize{$40$}}
			\end{overpic}
		\end{minipage}
		
	\end{tabular}
	\vspace{-0.2cm}
	
	\caption{
		\label{fig:opt}  We investigate the best possible matching achievable with our embedding. On the left, we report the error varying the embedding dimension, comparing to the standard LBO basis $\LBOBasis$ on the two settings S+F and F+S. Our basis always performs better. On the right, a qualitative example from F+S; the heatmap encodes the geodesic error of the correspondence.
		}

\end{figure*}

\section{Results}
\label{sec:results}

\subsection{Analysis of the learned embeddings}
We study the properties of our learned embedding, and we note that: i) It is always full rank, despite not being orthogonal (unlike the LBO basis, which is orthogonal by construction); ii) 
It is smooth. LBO eigenfunctions form the smoothest orthonormal basis.
In Figure \ref{fig:basis}, we depict the embedding dimensions as functions over the surfaces and report their Dirichlet energy. Even if we do not explicitly impose smoothness, our embedding is significantly smoother than LBO one (smaller $E_{D}$); this is possible since we do not impose orthogonality.

\paragraph{Basis dimension.} First, we compare our embedding $\OurBasis$ with the LBO one $\LBOBasis$. Moreover, we evaluate how the performance depends on the embedding dimension. We train $5$ embedding networks with a different number of output dimensions (i.e., $5$, $10$, $20$, $30$, and $40$) both on $S+F$ and $F+S$. On the test dataset, to focus only on the embedding properties, we compute for each embedding the optimal transformation between the $50$ pairs as done in Equation \ref{eq:defcopt}, and we estimate the correspondence solving the nearest neighbor search ($NN$) as $\Pi_{\mathcal{M}\mathcal{N}} = NN(\Phi_{\mathcal{M}} C^{opt}_{\mathcal{N}\mathcal{M}}, \Phi_{\mathcal{N}})$.
Even using the ground truth $\Pi^{gt}_{\mathcal{M}\mathcal{N}}$ to compute $C^{opt}_{\mathcal{N}\mathcal{M}}$, the embeddings may present some differences which are not linearly alignable. We measure the error generated by this loss of information. This choice makes our comparison independent of the estimation of the linear transformation.

The Table on the left of Figure \ref{fig:opt} contains the results. 
The first observation is that increasing the embedding size improves both $\OurBasis$ and $\LBOBasis$ representations. Secondly, we can compress the LBO representation to significantly smaller dimensions (i.e., with $5$ learned dimensions, we have the same quality of $20$ LBO basis; with $10$ learned, the same of $30$ LBO). But finally, and most importantly, our basis consistently outperforms the LBO basis. On the right of Figure \ref{fig:opt}, we also visualize the error of the two competitors. With small bases, the $\LBOBasis$ cannot distinguish between the back and front of the shape, and at higher dimensions, the error peak at the protrusions. These are classical errors for $\LBOBasis$, which suffers the front-back symmetry and the missing details on the tiny components such as arms and legs due to the low pass representation. Learned embedding distributes the error more homogeneously over the surface.

\setlength{\columnsep}{5pt}
\begin{wraptable}[4]{R}{0.2\linewidth}
	\footnotesize
	\vspace{-0.9cm}
    
    \begin{center}	    	
	\hspace{-1cm}

	\begin{tabular}{lr}
                            &  F+S \\ \hline
		$\OurBasis$ &  \textbf{1.58}  \\
		$\OurBasis_{HKS}$   &  1.81   \\ \hline
		$\LBOBasis$         &  2.05   \\

	\end{tabular}
    \end{center}

\end{wraptable}
\paragraph{Rotation invariance}
In shape matching, the rotation invariance is a desirable property. Our data augmentation is an effective technique to build resilience to rotations without giving up extrinsic information. 
To clarify this, we considered a pair of test shapes in the F+S setting. We fixed one of the two, applying  50 different random 3D rotations to the second. The error without rotations is $2.43$, while the mean error after the rotations is $2.52$ with a standard deviation of $0.11$. Moreover, in the inset Table, we input $50$ scales of HKS (a rotation-invariant descriptor) to train a $40$-dimensional embedding as done in \cite{sharp2020diffusion}.
Even if  $\OurBasis_{HKS}$ does not overcome $\OurBasis$ trained on coordinates, it is better than LBO, showing that our method could work with different input features.

\setlength{\columnsep}{5pt}
\begin{wraptable}[11]{R}{0.38\linewidth}
	\footnotesize
	\vspace{-0.8cm}
    
    \begin{center}	    	
	\hspace{-1cm}

	\begin{tabular}{@{}lrrr@{}}
		&  S+F    & F+S    & SMAL                  \\
		\hline
		$\OurBasis$      &  \textbf{1.58}    &  \textbf{1.52} & \textbf{1.8}  \\
		$\LBOBasis$                    &   2.05   &  1.97 & 4.3 \\
		L-Invariant            &   2.94   &   2.85 & 2.4 \\
		$L_{\ell1+E_D}$             &  13.18    &  5.72  & 14.9  \\
	\end{tabular}
    \end{center}
   \vspace{-0.2cm}

		\caption{\label{tab:losses}
	Quantitative comparison on shape matching, learning a $40$ dimensional embedding and computing the ground truth transformation $C^{opt}$. Each row is a different training.}

\end{wraptable}
\paragraph{Comparison of learned embeddings.}
We compare $\OurBasis$ and $\LBOBasis$ to learned embeddings that exploit: i) a different architecture (L-Invariant obtained from \cite{LIE2020}) and ii) and a different loss (with $L_{\ell1+E_D}$ and without $L_{\EmbNet}$).
Again, we adopt the ground truth linear transformation $C^{opt}$ highlighting the performance of each alternative in the optimal setting.
In Table \ref{tab:losses}, we report the results for the matching task. As expected, the embeddings learned with the supervised loss $L_{\EmbNet}$ perform better. However, we note that $L_{\ell1+E_D}$ preserves some matching capacity in the F+S setting, highlighting interesting relations between matching and general properties, which merit further explorations as future work. Finally, L-Invariant reveals that \cite{LIE2020} cannot overtake $\LBOBasis$ on meshes. 

\begin{table}
    \begin{minipage}{0.43\linewidth}
	\begin{center}
	\footnotesize
	\begin{tabular}{@{}lrrr@{}}
		&  S+F    & F+S    & SMAL                  \\
		\hline
		$\OurBasis$      &  1.58    &  1.52 & 1.8 \\
		$\OurBasis_{+L_{E_D}}$          & 1.29 & \textbf{1.34}  & 2.0 \\
		$\OurBasis_{+\Lo}$       & 15.35 & 23.10 & 19.0 \\
		$\OurBasis_{+\Low}$  &  \textbf{1.19} & 1.41 & \textbf{1.4}\\
		$\OurBasis_{+L_{E_D}+\Low}$ & \textbf{1.19} & \textbf{1.35} & 5.9\\
	\end{tabular}
	\end{center}
	\caption{\label{tab:losses_reg}
	Quantitative comparison on different regularizations, learning a $40$ dimensional embedding, and computing the ground truth transformation $C_{opt}$. Each row is a different training.}
    \end{minipage}
	\hspace{0.3cm}
    \begin{minipage}{0.54\linewidth}
	\begin{center}
		\footnotesize
	\begin{tabular}{@{}lrrrrr@{}}
		&  $\Snob$+F  &  $\Snob$+$\Fnob$    & $\Fnob$+S & $\Fnob$+$\Snob$   & \test{S+F}        \\
		\hline
		$\LBOBasis$&                               2.05 & 1.88 & 1.97 & 2.04 & 2.05\\
		$\OurBasis$&               1.12 & \textbf{0.76} & \textbf{1.24} & \textbf{1.27} & \textbf{1.07}\\
		$\OurBasis_{+\Ls}$&         \textbf{1.10} & \textbf{0.77} & \textbf{1.23} & \textbf{1.27} & \textbf{1.07}\\
		$\OurBasis_{+\Low}$& 1.60 & 1.61 & 1.50 & 1.55 & 1.14  \\
		$\OurBasis_{+\Ls+\Low}$ & 1.77 & 1.76 & 1.58 & 1.61 & 1.19\\
	\end{tabular}
	\vspace{0.5cm}
	 
	\caption{\label{tab:nonbj}
	Quantitative comparison on shape matching, learning a $40$ dimensional embedding, and computing the ground truth transformation $C_{opt}$. Each row is a different training.}
    \end{center}
    \end{minipage}
\end{table}

\paragraph{Regularizations in the bijective scenario.} Noticing the potential of the learned embedding, we decided to explore additional regularizations during the training, starting from the bijective setting. Table \ref{tab:losses_reg} shows the comparisons obtained combining $L_{\EmbNet}$ with one or more losses from Section\ref{sec:method}. Promoting the smoothness ($\OurBasis_{\ell+L_{E_D}}$) produces a slight improvement or keeps stable results both with and without $\Lo$. Our analysis anticipated this result since $L_{\EmbNet}$ produces smoothness as a good property for the matching. Including $\Lo$ instead is disruptive due to its magnitude, which strongly bounds the network to favour the bijectivity induced by $\Lo$ over good matching. We thus scale $\Lo$ to a magnitude similar to $\Ls$. With $\OurBasis_{+\Low}$, the network obtained a significant gain. We highlight that SMAL is the only dataset where smoothness seems slightly dangerous. We attribute this drop to the discontinuous self-intersections and artifacts present on the muzzle of animals. 

\paragraph{Regularizations in challenging scenarios.} Given the previous results, we verify our intuitions considering challenging contexts without bijection. Unlike previous works (that considered remeshed datasets where shapes have a similar density), we consider configurations with a significant change in the number of vertices. In Table \ref{tab:nonbj}, we train and test on different scenarios as described in Section \ref{sec:settings}. Once again, the smoothness does not change the performance. Adding orthogonality confirms our hypothesis: promoting bijectivity when this is not present worsen the results. Finally, in the last column, even if the shapes inside each dataset are in bijection, orthogonality does not help in the presence of huge density variations.



\subsection{Applicative test}

\begin{figure*}[!t]
\begin{tabular}{cc}
\hspace{-0.3cm}
\begin{minipage}{0.29\linewidth}
    \footnotesize
	\begin{tabular}{@{}lrr@{}}
	& \multicolumn{2}{c}{SMAL} \\
		          & $\Phi C$   & $\Phi C^{opt}$       \\
		\hline
		$Our^{\EmbNet}$                & 5.67 & 1.8\\
		$Our^{\EmbNet}_{+\Ls}$         &   11.0   & 2.0\\
		$Our^{\EmbNet}_{+\Low}$         &  \textbf{3.9} & 1.4    \\
		$Our^{\EmbNet}_{+\Ls+\Low}$         &   11.5   & 5.9\\ \hline
		Universal          &  5.91               & 4.3\\
		$\DiffNet$         &  7.94               & 4.3\\
		$GFM_{KP}$ \cite{donati2020deep} &     8.00 & 4.3\\ 
        Deep Shells \cite{eisenberger2020deep}	& 15.23 & 4.3\\
        L-Invariant \cite{LIE2020} &	21.10    & 4.3\\
        
	\end{tabular}
	\vspace{-0.2cm}
	
\end{minipage}
&
\hspace{0.3cm}

\begin{minipage}{0.6\linewidth}
	\scriptsize
	\vspace{0.2cm}
	\begin{overpic}[trim=0cm 0cm 0cm 0cm,clip, width=1.04\linewidth]{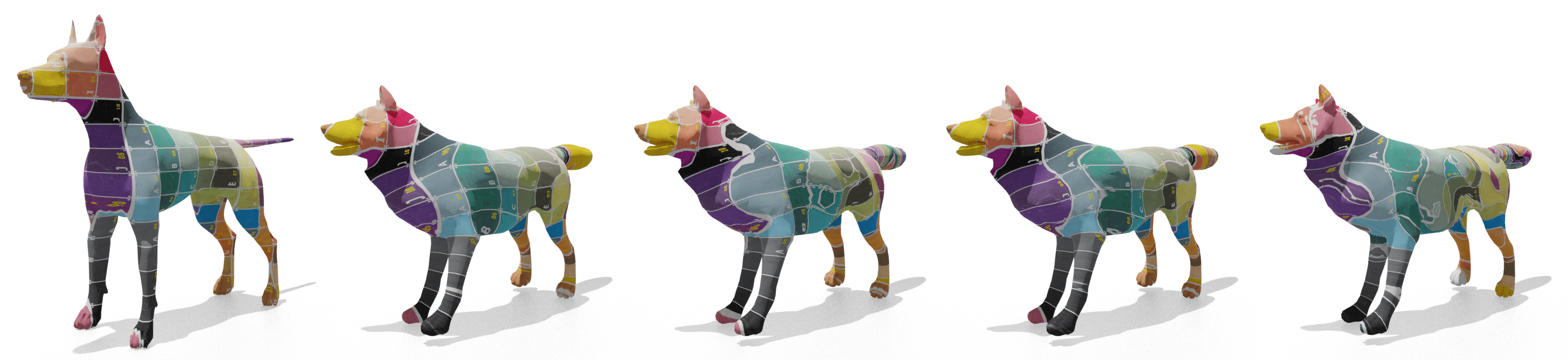}
		\put(8,-2.5){$\M$}
		\put(27,-2.5){Our}
		\put(44,-2.5){$GFM_{DN}$}
		\put(48,-6.0){\cite{donati2020deep}}
		\put(64,-2.5){Universal}
		\put(82,-2.5){L-Invariant} \put(87.5,-6.0){\cite{LIE2020}}
	\end{overpic}
	\vspace{-0.3cm}

\hspace{-0.2cm}
\end{minipage}

\end{tabular}

	\caption{\label{fig:tosca} On the left, a quantitative comparison for non-isometric shape matching on SMAL dataset; the first column is learning also the descriptor, while the second is the theoretically best possible matching with that embedding using the optimal transformation $C^{opt}$. On the right, a pair of animals from TOSCA \cite{TOSCA} dataset, showing our generalization capability.}
\end{figure*}
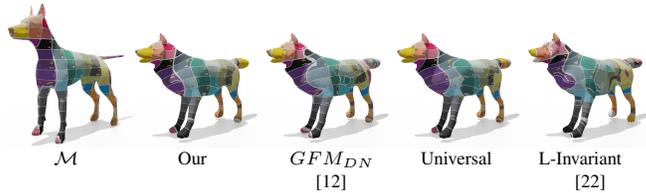

\paragraph{Learning descriptors}
Many previous works that we discussed in Section \ref{sec:related}, validate their matching performance on the S+F and F+S settings. We report in the Supplementary Material the results of our approach on them, learning also the descriptors. Here we focus on more exciting challenges: non-isometric shapes from SMAL, in which the considered animals may have a significant difference in their surface metric. In the Table on the left of Figure \ref{fig:tosca}, we report the results. 

\paragraph{Competitors}
We considered four competitors in our analysis. \emph{Universal} shares our same architecture but it does not foster a linear transformation to align them and is optimized with a different loss:
\begin{equation}
    L_{U}(\Phi_{\M}^{uni}, \Phi_{\N}^{uni}) = \sum \|\Phi_{\M}^{uni} -  \Pi^{gt}_{\M\N} \Phi_{\N}^{uni} \|_2^2\,.
\end{equation}
$L_{U}$ promotes universal signatures for the points in a similar fashion as the descriptor-based methods described in Section~\ref{sec:related}. This comparison is essential to highlight the merits of considering linear alignable embeddings.
%
The other main competitor is \cite{sharp2020diffusion}, referred to as $\DiffNet$. As done in our pipeline, this method adopts the same architecture to learn the features, but it exploits the $\LBOBasis$ instead of $\OurBasis$. This comparison highlights the contributions of the learned embedding with respect to $\LBOBasis$.
We will also consider two other methods to align $\LBOBasis$: i) $GFM_{KP}$ \cite{donati2020deep}, that uses a KPConv backbone \cite{Thomas_2019_ICCV}, and ii) \emph{Deep Shells} \cite{eisenberger2020deep} that relies on a registration approach. For completeness, we also report the results of \emph{L-Invariant} with its PointNet network \cite{LIE2020}. In the experiments, we measure the error as the geodesic distance between the predicted match and the ground truth, and we report the average over all the points. 
While our analysis mainly focuses on theoretical properties of learned embeddings, we would emphasize contexts in which they show significant applicability. Not only we can learn descriptors that surpass state-of-the-art methods, but we also obtain better performance than the \emph{theoretically best possible results with LBO}. Further preliminary evidence on descriptor learning, matching refinement, and functional reconstruction can be found in Supplementary Material. 

We notice that, without any further regularization, we can recover good descriptors even for $\OurBasis$. Once again, we observe that smoothness is not particularly favorable in this context. Finally, $Our^{\EmbNet}_{+\Low}$ gives rise to the most exciting result, confirming our theoretical and empirical insights and outperforming the best possible performance achievable by LBO (that is 4.3, from Table \ref{tab:losses}).

\paragraph{Computational specifications}
We use an i9-9820X Intel processor at 3.31GHz and an RTX 2080 Ti GPU card for the experiments. Embedding training requires $\sim$ 180min for the F+S setting and 210min for S+F, regardless of the number of basis functions. 


\section{Conclusions}
\label{sec:conclusion}

\paragraph{Insights.} The observed saturation on the basis sets of increasing dimensions is caused by different reasons. Including more $\LBOBasis$ basis functions introduces instability due to high frequencies that are hardly alignable linearly. Instead, $\OurBasis$, being a learning method, is prone to overfitting if the representation is too rich. 
Smoothness seems an emerging natural property from matching. Intuitively, a linear alignment is simpler to achieve if there are no outliers or drastic changes; also, in our cases, the correspondence is a smooth function over the surface. However, if the data present discontinuities, the smoothness becomes dangerous. Orthogonality seems advantageous in contexts where bijectivity is a good prior. On the other hand, it requires proper weighting and could harm the results even if training and tests are not bijective. We believe this latter is a consequence of the global nature of orthogonality, which limits its adaptation if shapes change their number of vertices. We highlight that fully intrinsic approaches are rotation invariant but do not distinguish symmetries and may unrecoverably degrade the performance. Finally, our non-isometric and applicative domain results highlight the importance of having a proper basis to further push the matching results.


\paragraph{Limitations.} While our work expands the understanding of learned embeddings, several questions remain. Investigating other classes of transformations different from the linear one is still entirely open. Other challenges like clutter and partial-to-partial matching (touched on in the Supplementary Materials) may require sophisticated strategies. Finally, cases in which the correspondence between two shapes is highly discontinuous would open to a completely different and intriguing analysis.

\paragraph{Future directions.} In recent years, several methods have relied on refining the matching using ZoomOut \cite{melzi2019zoomout}. This is a significantly important aspect for real applications, and it may seem a drawback of learned embedding of which dimension must be decided at training time. In the Supplementary Materials, we show that thanks to their smoothness, the learned basis can be augmented by ZoomOut, refining the matching by including the LBO eigenfunctions at increasing frequencies. These results are novel and exciting, but even more importantly, they pave the way for further exploration of possible refinement techniques designed explicitly for learned embeddings. Moreover, we focus on the shape matching task, but the learned embeddings can target other applications. In the Supplementary Materials, we collect some coordinate function transfer results, and we foster other possible tasks as fascinating future work.  

Our work shows a series of promising properties. We wish to motivate the community to foster research in this direction. Pointing out that smoothness seems a superior property to orthogonality in some contexts sheds new light on the study of shape embedding for matching, which for many years focused in a contrary direction \cite{LMH,hamiltonian}. Generalization to challenging scenarios is appealing and highlights the importance of proper embedding instead of refining existing ones. Our joint analysis of learned and axiomatic embedding opens to look for other properties to impose and a broad set of different applications.

\paragraph*{Acknowledgements}
This work is supported by the ERC Starting Grant No. 802554 (SPECGEO), the SAPIENZA BE-FOR-ERC 2020 Grant (NONLINFMAPS), KAUST OSR Award No. CRG-2017-3426, the ERC Starting Grant No. 758800 (EXPROTEA) and the ANR AI Chair AIGRETTE. 

\medskip
\bibliographystyle{splcs04}
\bibliography{eg_bib}






\appendix

\section{Appendix}
\label{sec:appendixA}

Given a network $\mathcal{N}$ that takes a shape $S$ and produces a $d$-dimensional embedding for every vertex on $S$. Stacking these values into a matrix (where rows correspond to vertices) we get a \textit{basis matrix} $\Phi_{S}$.

Now, suppose the network $\mathcal{N}$ is trained so that for any input shape $S$, the basis matrix $\Phi_S = \mathcal{N}(S)$ should be approximately orthonormal: $\Phi_S^{\top} \Phi_S \approx Id$. We will call such a network $\mathcal{N}$ an \emph{orthonormal} feature network.

In a simpler scenario, suppose that we use the trained network to compute dense correspondences across shape pairs using nearest neighbor search in embedding space. Thus, given two shapes $S_{1}, S_{2}$, we first compute the basis matrices $\Phi_1 = \mathcal{N}(S_{1})$ and $\Phi_2 = \mathcal{N}(S_2)$. We then compute a map by solving the problem:
\begin{align}
\label{eq:densemap}
    \Pi_{S_1 S_2} = \argmin_{\Pi \in \mathcal{P}} \|  \Phi_1 - \Pi \Phi_2 \|^2.
\end{align}
Here $\mathcal{P}$ is the space of all binary matrices that represent point-to-point maps: i.e., binary matrices with exactly one 1 per row, and we use the Frobenius norm. Note that the optimization problem in Eq.\eqref{eq:densemap} is row separable, meaning that it can be solved via nearest neighbor search in the $L_2$ sense.

\begin{lemma}
\label{lemma:ortho_g}
If the basis matrices are \emph{exactly} orthonormal and the residual when solving for the map in Eq.~\eqref{eq:densemap} is zero, then the optimal map $\Pi = \Pi_{S_1 S_2}$ must satisfy $\Phi_2^\top \Pi^\top \Pi \Phi_2 = Id$.
\end{lemma}
\begin{proof}
By assumption we have $\Phi_1 = \Pi \Phi_2$. Moreover since $\Phi_1^\top \Phi_1 = Id$, left-multiplying the first equation by $\Phi_1^\top$, we get: $\Phi_1^\top \Phi_1 = \Phi_1^\top \Pi \Phi_2$, so that $\Phi_1^\top \Pi \Phi_2 = Id$. Using again the fact that $\Phi_1 = \Pi \Phi_2$ we get $\Phi_2^T \Pi^\top \Pi \Phi_2 = Id$.
\end{proof}

\begin{lemma}
\label{lemma:full_d}
Under the assumptions of Lemma \ref{lemma:ortho_g}, if, moreover, the dimensionality $d$ of the basis matrices equals $n$, the number of points on $S_1$, then the optimal map $\Pi = \Pi_{S_1 S_2}$ must be orthonormal, and thus represent a bijective map.
\end{lemma}
\begin{proof}
If $d=n$ then $\Phi_2$ is a square matrix, and since it is orthonormal, it must have an inverse $\Phi_2^{-1} = \Phi_2^\top$. Using this fact and applying Lemma \ref{lemma:ortho_g} we get $\Pi^\top \Pi = Id$. Since $\Pi$ is, by assumption a binary matrix, it must therefore be doubly stochastic and represent a bijective map.
\end{proof}

\begin{lemma}
\label{lemma:reduced_d}
More generally, for an arbitrary dimension $d$ of the basis matrix, suppose that there is a subset of vertices $S \subseteq S_{2}$ such that the indicator function $I_{S}$ lies within the span of the basis matrix $\Phi_2$. I.e., $I_{S} = \Phi_2 \mathbf{a}$ for some vector of coefficients $\mathbf{a}$. Then, again under the assumptions of Lemma \ref{lemma:ortho_g}, we have $\|\Pi I_{S} \| = \|I_{S}\|$. Thus the map $\Pi$ must preserve the cardinality of the set $I_{S}$, or, in other words, $\Pi$ \textbf{must be injective} when restricted to $S$.
\end{lemma}
\begin{proof}
We have $\|I_{S}\| = \mathbf{a}^T \Phi_2^\top \Phi_2 \mathbf{a} =  \mathbf{a}^\top  \mathbf{a}$. On the other hand, $\| \Pi I_{S} \| = \mathbf{a}^\top \Phi_2^\top \Pi^\top \Pi \Phi_2 \mathbf{a} = \mathbf{a}^\top  \mathbf{a}$ using Lemma \ref{lemma:ortho_g}.
\end{proof}

\begin{lemma}
\label{lemma:reduced_d_interp}
In particular, if the feature matrix $\Phi_2$ contains the constant function, regardless of the dimensionality, then under the assumptions of Lemma~\ref{lemma:reduced_d}, $\Pi$ must be injective on the entire shape. Moreover, if size of $\Phi_1,\Phi_2$ is the same, then the map must be bijective.
\end{lemma}
\begin{proof}
The first part follows directly from Lemma \ref{lemma:reduced_d}. For the second part, observe that if $\|\Pi I_{S}\| = \|I_{S}\|$ then, if $I_{S}$ is the indicator of the entire shape, the cardinality of $\Pi I_{S}$ is the same as the number of rows in $\Phi_1$, which means that the map is also surjective, and thus, must be bijective.
\end{proof}

\clearpage 
\newpage

\appendix
\renewcommand*{\thesection}{\arabic{section}}

\null
  \vbox{%
    \hsize\textwidth
    \linewidth\hsize
    \vskip 0.1in
  \hrule height 4px
  \vskip 0.25in
  \vskip -\parskip%
    \centering
    {\LARGE\bf Supplementary Materials\par}
  \vskip 0.29in
  \vskip -\parskip
  \hrule height 1px
  \vskip 0.09in%
    \vskip 0.2in
  }
  
\begin{abstract}
Here we list additional material as support to the main manuscript. We provide more details about the training processes and hyperparameters choice. We also offer an analysis on the relationship between the cardinality of basis and descriptors, quantitative results in isometric cases, and qualitative results both on humans and non-isometric animals. We describe more in detail the partiality setting, and finally, we include some more examples of ZoomOut process.
\end{abstract}
\section{Training Details}
\label{sec:training}

We compose the Diffusion networks with $12$ Diffusion Blocks in all cases. In the diffusion process, we use $128$ eigenvalues and eigenvectors. We use an ADAM optimizer with an initial learning rate of $10e^{-4}$ and a Cosine Annealing Warm Restarts scheduler. Due to computational reasons, we use batch size $2$ and accumulate the gradient for $4$ batches to simulate a batch size of $8$. We set the maximum number of epochs to $1200$. We will release our data and method implementation. 

\paragraph{Datasets}. 
We consider two human datasets: the $100$ shapes of FAUST \cite{FAUST}, a dataset of ten different subjects in ten different poses, and the $71$ shapes of SCAPE \cite{anguelov2005scape}, a single subject in $71$ different poses. Both of them are used with their original connectivity. At test time, we evaluate on $50$ random pairs of the testing dataset. 

SMAL dataset comprises different animal groups (big cats, canines, ovine, bovine, and hippos), presenting significant non-isometries. The training set is composed of 25 shapes of different species and poses, that is definitely small if compared to the ones generally adopted. The test set is composed of 300 pairs of 25 shapes unseen at training time. 

\subsection{Loss stability}
\begin{figure*}
    \centering
    \begin{tabular}{lr}
    \begin{minipage}{0.5\linewidth}
			\begin{overpic}[trim=0cm 0cm 0cm 0cm,clip, width=0.95\linewidth]{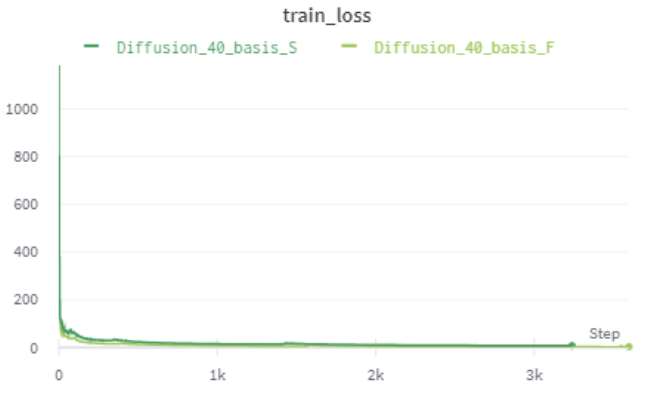}
			\end{overpic}
    \end{minipage}
         &      
    \begin{minipage}{0.5\linewidth}
			\begin{overpic}[trim=0cm 0cm 0cm 0cm,clip, width=0.95\linewidth]{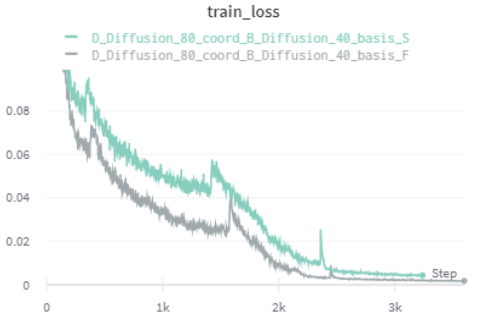}
			\end{overpic}
    \end{minipage}
         \\
         \multicolumn{2}{c}{
        \begin{minipage}{0.5\linewidth}
			\begin{overpic}[trim=0cm 0cm 0cm 0cm,clip, width=0.95\linewidth]{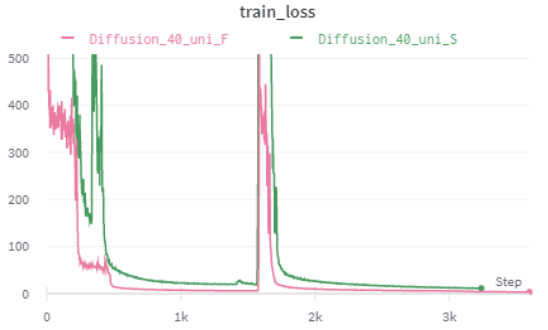}
			\end{overpic}
    \end{minipage}
    }
    \end{tabular}
    \caption{Training losses of our embedding (top left), features (top right) and Universal embedding formulation (bottom).}
    \label{fig:losses}
\end{figure*}

In Figure \ref{fig:losses}, we report the behavior of training loss when we train our basis, our descriptor, and the universal embedding in the two settings of S+F and F+S. Both our networks show stability. We attribute the peaks to the Cosine Annealing scheduler restart, which increases the learning rate to escape from local minima. For this reason, we think our solution spaces are more regular than the one of the universal embedding.

\clearpage 
\section{Function representation and coordinates as proxies}

\label{sec:func}
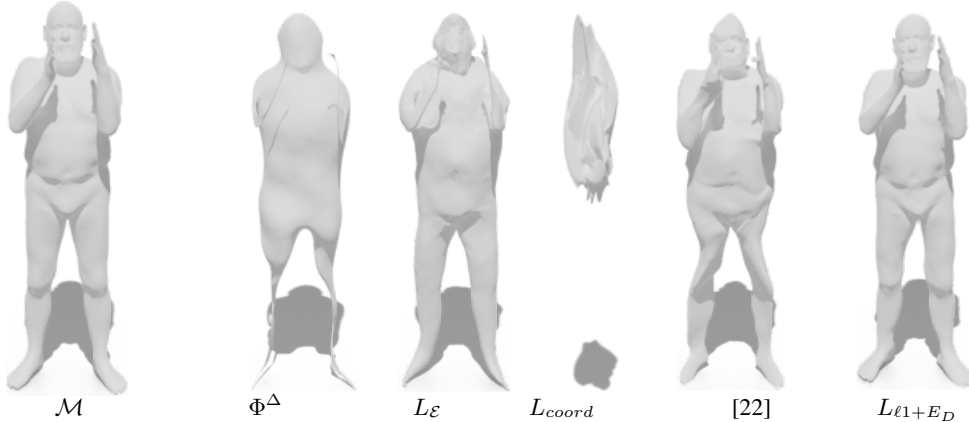
\begin{figure}[!t]
	\centering
	\footnotesize
	
	\begin{overpic}[trim=0cm 0cm 0cm 0cm,clip, width=0.92\linewidth]{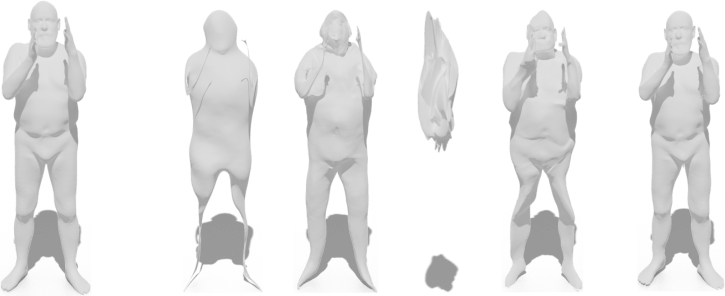}
		\put(5,-2){$\M$}
		\put(25,-2){$\LBOBasis$ }
		\put(42,-2){$L_{\EmbNet}$}
		\put(54,-2){$L_{coord}$}
		\put(75, -2){\cite{LIE2020}}
		\put(90, -2){$L_{\ell1+E_D}$}
	\end{overpic}
	\vspace{0.2cm}
	
	\caption{\label{fig:coord} Coordinate reconstruction with different basis. Optimizing only the coordinates brings to a dramatical overfitting. LBO causes loss of information on the protrusions. Optimizing for the matching produce sharper results, while considering smoothness and sparsity may produce better detail recover.}
\hspace{-0.2cm}

\end{figure}

\begin{wraptable}[10]{R}{0.36\linewidth}
	\footnotesize
	\hspace{1.6cm}
	\vspace{-0.6cm}
	
	    	\begin{minipage}{\linewidth}
	    	\hspace{-1.8cm}
	    	
			\begin{tabular}{lrr}
		                &        F+S       & S+F \\ \hline
		$\Phi_{\EmbNet}$     & 2.86    &  2.26    \\
		$\Phi_{\ell1+E_D}$      &  \textbf{1.07}   &   \textbf{0.89}   \\ 
		$\Phi_{Coord}$     &  47.58  &   45.90  \\
		$\LBOBasis$         &  2.96    &  3.06\\
        L-Invariant \cite{LIE2020}     &   3.62   & 1.91   \\
		L-Invariant $L_{\ell1+E_D}$ & 25.35 & 44.40 
			\end{tabular}
	\vspace{-0.3cm}
	\hspace{1cm}

	\caption{\label{tab:coord} Results in coordinates representation.}
	\end{minipage}
\end{wraptable}
Until now, the functional perspective has been carried only by the Functional Maps paradigm. How learned embeddings can represent functions was left unexplored by \cite{LIE2020}, but we think it is an important aspect, given the number of tools and analogies that it suggests. We focus on coordinates as a triplet of functions defined over the vertices. The interest in how the basis can represent such functions is multiple. Coordinates evolve smoothly across a shape, particularly in the human domain; hence, they are good examples of a continuous global function. Representing the coordinates of each point can be visualized, and it gives an interpretation of the loss of information. Finally, and importantly, reconstructing the coordinates means that the basis can represent the points on the surface. In this case, we considered three different trainings of $\OurBasis$. In the first one, we used the $L_{\EmbNet}$ defined in Eq. 4 of the main manuscript. Then, we trained using $L_{\ell1+E_D}$ presented in Equations 7 and 8 of the paper. To highlight the benefit of our mesh approach, we also compare \cite{LIE2020} trained on $L_{\EmbNet}$ and $L_{\ell1+E_D}$. Finally, we consider the case in which we train the features to represent the coordinates. We perform this training considering the following loss:
\begin{equation}
    L_{coord}(\OurBasis_{\M}) = \sum \|\OurBasis_{\M} \OurBasis_{\M}^{\dagger} X_{\M} - X_{\M}\|_2^2,
    \label{eq:coord}
\end{equation}
which quantifies the error in the reconstruction of the coordinates functions $X_{\M}$, provided by the learned basis.
In Table \ref{tab:coord} we report the results of our experiment on F+S and S+F settings. Error is expressed as Equation \ref{eq:coord}. A qualitative example is reported in Figure \ref{fig:coord}.

\paragraph{Insights.} 
Interestingly, learning for matching generates good bases also for function representation. The geometry of the shape is significantly more inflated than in the LBO case, in which the protrusions collapsed. $L_{\ell1+E_D}$ also gave rise to an interesting result, in which we obtain a precise characterization, even in the finer details. This result shows that a functional basis that produces a good representation of the vertices does not imply an improved matching capability. The ad-hoc training with $L_{coord}$ is the worst result, probably due to a lack of generalization: learning to reproduce specific functions for a small set of shapes may cause a quick overfit. Giving structure to the embeddings seems promising in obtaining more general tools for shape analysis. We finally highlight that matching loss for L-Invariant trained with $L_{\EmbNet}$ produces worst results on F+S settings but a better result in S+F where the poses seen at training time are more comprehensive than the ones observed at test time. The complete absence of intrinsic information in \cite{LIE2020} makes it harder to generalize on new poses. Finally, L-Invariant cannot be trained with $L_{\ell1+E_D}$; we believe this is a strong motivation for our approach.

\begin{wraptable}[4]{R}{0.35\linewidth}
	\footnotesize
	\vspace{-1.4cm}
	    	
			\begin{tabular}{lrrr}
				                    &  0.0    & 0.4       & 0.8 \\ \hline
				$\OurBasis C_{opt}$ &  \textbf{1.58}   & \textbf{6.8}     &  \textbf{10.5}\\
				$\LBOBasis C_{opt}$ &  2.05   &  7.3             &   10.6   \\ \hline
				Our$^\EmbNet$                 &  3.80       & 8.4              &   12.0      \\
				$\DiffNet$       &  \textbf{2.90} &    \textbf{8.2}    &  \textbf{11.9}\\

			\end{tabular}
			\caption{\label{tab:partiality} Results with different partialities in S+F setting, training on full shapes.}
\end{wraptable}
\section{Partiality.}
\label{sec:partiality}

		\begin{figure*}
		\centering 
			\begin{overpic}[trim=0cm 0cm 0cm 0cm,clip, width=0.7\linewidth]{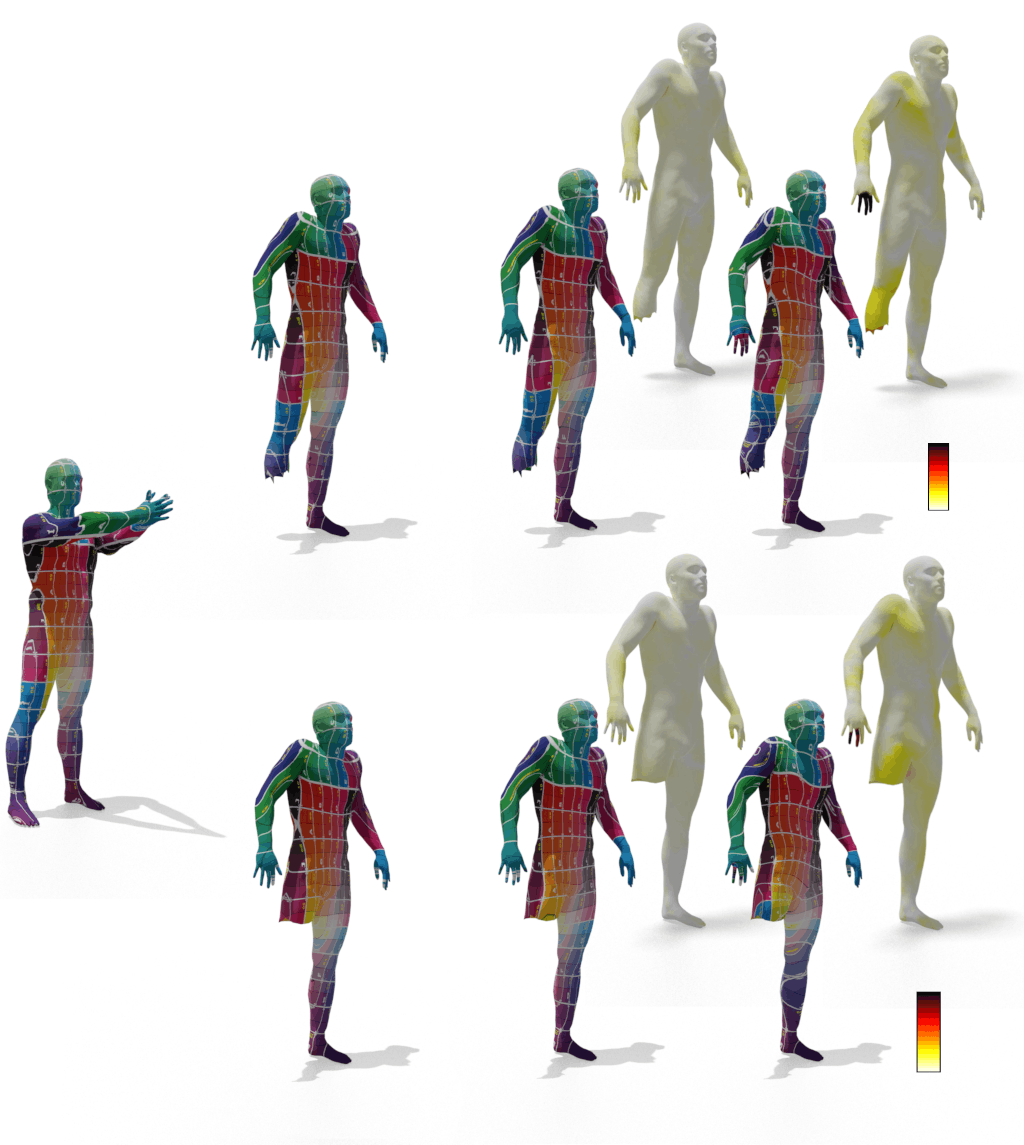}
			\put(5,25){\scriptsize{$\M$}}
			\put(28,47){\scriptsize{GT}}
			\put(48,47){\scriptsize{Our}}
			\put(66,47){\scriptsize{$\DiffNet$}}
			
			\put(18,75){\scriptsize{$0.4$}}
			\put(18,20){\scriptsize{$0.8$}	}		

			\put(85,55){\scriptsize{$0$}	}
			\put(85,61){\scriptsize{max}	}

			\put(85,6.5){\scriptsize{$0$}	}
			\put(85,12.5){\scriptsize{max}	}
			
			\put(50,95){\scriptsize{4.58}}
			\put(50,5){\scriptsize{4.57}}
			\put(70,95){\scriptsize{11.04}}
			\put(70,5){\scriptsize{8.98}}
			
			\end{overpic}
			\caption{\label{fig:partial1} An example of partiality involved in our experiments. On the top, the $0.2$ setting. On the bottom, the $0.4$ one. Comparison between Our and $\DiffNet$. Near to each case we report the avarage geodesic error over the surface.}
		\end{figure*}

		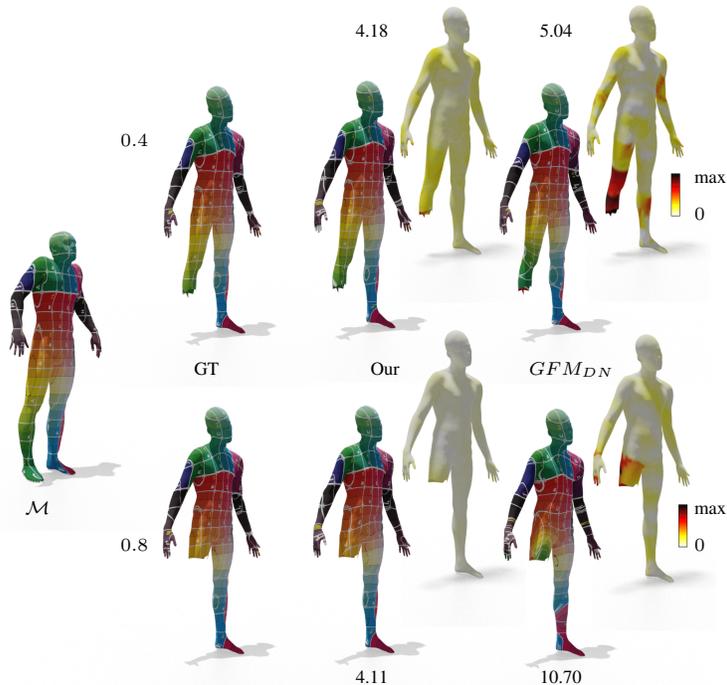
\begin{figure*}
		\centering 
			\begin{overpic}[trim=0cm 0cm 0cm 0cm,clip, width=0.7\linewidth]{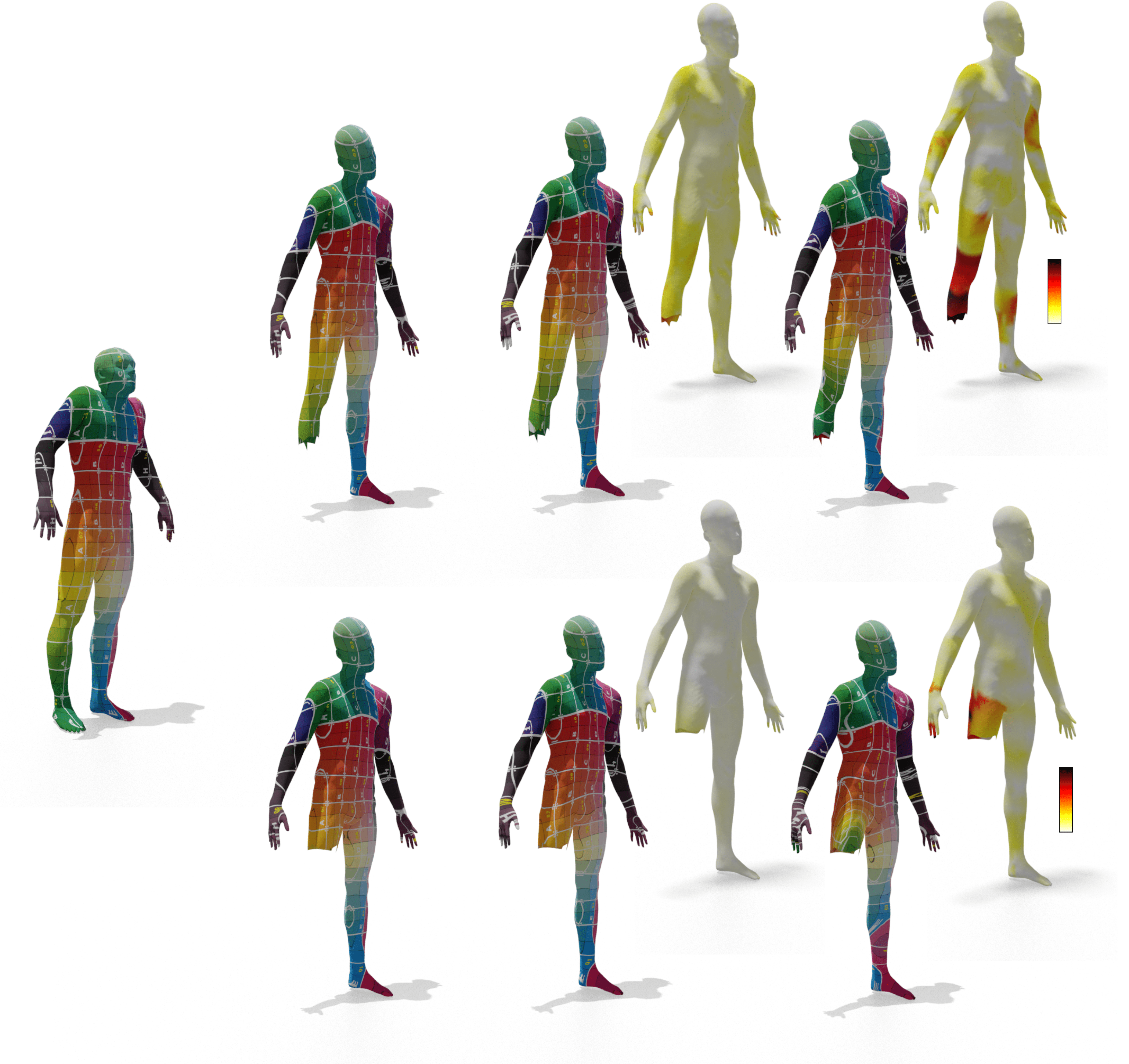}
			\put(5,25){\scriptsize{$\M$}}
			\put(28,44){\scriptsize{GT}}
			\put(52,44){\scriptsize{Our}}
			\put(73.5,44){\scriptsize{$\DiffNet$}}
			
			\put(18,75){\scriptsize{$0.4$}}
			\put(18,20){\scriptsize{$0.8$}	}		

			\put(96,20){\scriptsize{$0$}	}
			\put(96,25){\scriptsize{max}	}
			
			\put(96,65){\scriptsize{$0$}	}
			\put(96,70){\scriptsize{max}	}
			
			\put(50,90){\scriptsize{4.18}}
			\put(50,2){\scriptsize{4.11}}
			\put(75,90){\scriptsize{5.04}}
			\put(75,2){\scriptsize{10.70}}

			\end{overpic}
			\caption{\label{fig:partial2} An example of partiality involved in our experiments. On the top, the $0.2$ setting. On the bottom, the $0.4$ one. Comparison between Our and $\DiffNet$. Near to each case we report the avarage geodesic error over the surface.}
		\end{figure*}
To test the resilience of learned features to missing geometry, we used $\OurBasis$ trained on the {\em full} shapes of SCAPE (i.e., without any data augmentation with partial shapes).
We constructed our dataset as follows:
\begin{enumerate}
    \item We consider the 100 FAUST shapes.
    \item For each shape, we take a landmark on the right foot.
    \item We remove all surface that is within a certain geodesic distance to the landmark. We consider two different ranges: $0.4$ and $0.8$.
\end{enumerate}
Then, we kept the $50$ pairs considered in the other experiments, substituting the source shape with the partial one. Hence, we look for a point on the complete shape for each point of the partial shape. Notice that the partial and the complete are different subjects in different poses. We want to remark that partiality has not to be seen at training time.
We argue that this dataset is more interesting from an applicative perspective compared to SHREC16 \cite{cosmo2016shrec}: the latter has been designed to analyze some specific theoretical properties of LBO basis. Instead, we are not aware of any other work that tackled the problem of missing limbs in human shape matching, also providing a more fair representation of different human beings. We show the results in Table \ref{tab:partiality}; in the first two rows, we consider the optimal matching, while in the last two, we consider the one with the features obtained by $\FeatNet$. 
In Figures \ref{fig:partial1} and \ref{fig:partial2} we depict two qualitative examples. On the left, the full shape. The ground-truth, our and $\DiffNet$ matchings in the $0.4$ setting, with texture transfer and error on the top right. On the bottom right, the same for the $0.8$ setting. The LBO bases are unstable, and the results of matching between $0.4$ and $0.8$ settings vary significantly even in regions far from the missing part (e.g., the right arm).

\begin{table*}[!h]
	\centering
	\ra{1}
	\begin{tabular}{@{}rrrrrcrrrr@{}}\toprule
		
		& \multicolumn{4}{c}{S+F}   &    \phantom{abc}                   & \multicolumn{4}{c}{F+S}                          \\
		\cmidrule{2-5} \cmidrule{7-10} 
		& \multicolumn{4}{c}{\textbf{\#Features}} & &\multicolumn{4}{c}{\textbf{\#Features}} \\
		
		& 20    & 40    & 80                  & 120    &    & 20         & 40    & 80                  & 120   \\
		\hline 
		5 \textbf{Basis}  & 34.75 & 30.80 & { 18.29}         & 29.81   &   & 15.28      & 13.14 & 10.61               & 15.56 \\
		10 \textbf{Basis} & 11.01 & 19.88 & {6.85}          & 15.03     & & {8.83} & 18.92 & 16.30               & 12.90 \\
		20 \textbf{Basis} & 32.07 & 9.83  & {4.67}          & 10.77    &  & 40.62      & 12.72 & {\textbf{7.68}} & 12.78 \\
		30 \textbf{Basis} & 14.60 & 13.18 & 3.98                & {3.46}& & {8.34} & 29.71 & 8.87                & 9.82  \\
		40 \textbf{Basis} & 7.61  & 29.66 & {\textbf{3.77}} & 4.63      & & 12.68      & 49.85 & {8.76}          & 11.49 \\
		\bottomrule
	\end{tabular}
	\caption{\label{tab:abl} Analysis with different number of basis embedding dimensions and features.}
\end{table*}
\section{Matching}

\label{Sec:matching}

\subsection{Basis and Descriptors relation}

Given that \cite{LIE2020} does not provide any insight on the relation between the basis and descriptors, we investigated this relation. In Table \ref{tab:abl}, we performed an extensive analysis with different combinations of embedding dimensions and a number of features.
Each entry of Table \ref{tab:abl} corresponds to a separate training. Concerning the dimensionality of the embedding, we notice that coherently with Figure 3 of the main manuscript, an increasing number of basis functions does not always produce better results. For example, in the F+S case, $40$ basis functions tend to be hardly alignable. We believe this is mainly because the SCAPE dataset contains a wide variety of poses. Hence, overfitting FAUST by using a larger representation would make it difficult to generalize at test time.

Our analysis also gives a good rule of thumb for features: a number bounded between $\times2$ and $\times3$ seems the one producing the most stable results. We consider this number reasonable to overcome the noise and in line with the literature~\cite{abid2017linear}.
This analysis sheds light on the role of the two components of the matching pipeline; the embedding should provide a structure shared across the objects to be easily alignable. The features help to identify such structure. 
A complex structure is not desirable, while sometimes redundant information in the features can help to identify the correct transformation.

Given such analysis, we decide to keep 40 dimensions for the basis and 80 for the features for all our experiments. This choice provides coherence in the results and good performance in all the settings.

\begin{wraptable}[6]{R}{0.25\linewidth}
	\footnotesize
	\vspace{-1.4cm}
    \centering
	\begin{tabular}{@{}lrrr@{}}
		&  S+F                       & F+S                     \\
		\hline
		Our$^\EmbNet$             & 3.8          & \textbf{8.8}   \\ 
		$\DiffNet$                    & \textbf{2.9}  & 10.2     \\
		Universal                     &  4.5          & 12.4                  
	\end{tabular}
	
	\caption{\label{tab:matching}
	Quantitative comparison on shape matching.}
\end{wraptable}
\subsection{Near-Isometric matching}
We tested our capability of matching near-isometric shapes using the learned embedding and features. In Table \ref{tab:matching}, we reported a comparison of our embeddings and main competitors in the S+F and F+S settings. For all the experiments, we kept a $40$-dimensional embedding and $80$. As can be seen, we perform better in F+S settings where generalization is relevant to obtain good results.

\begin{figure*}
\centering

		\begin{overpic}[trim=0cm 0cm 0cm 0cm,clip, width=0.90\linewidth]{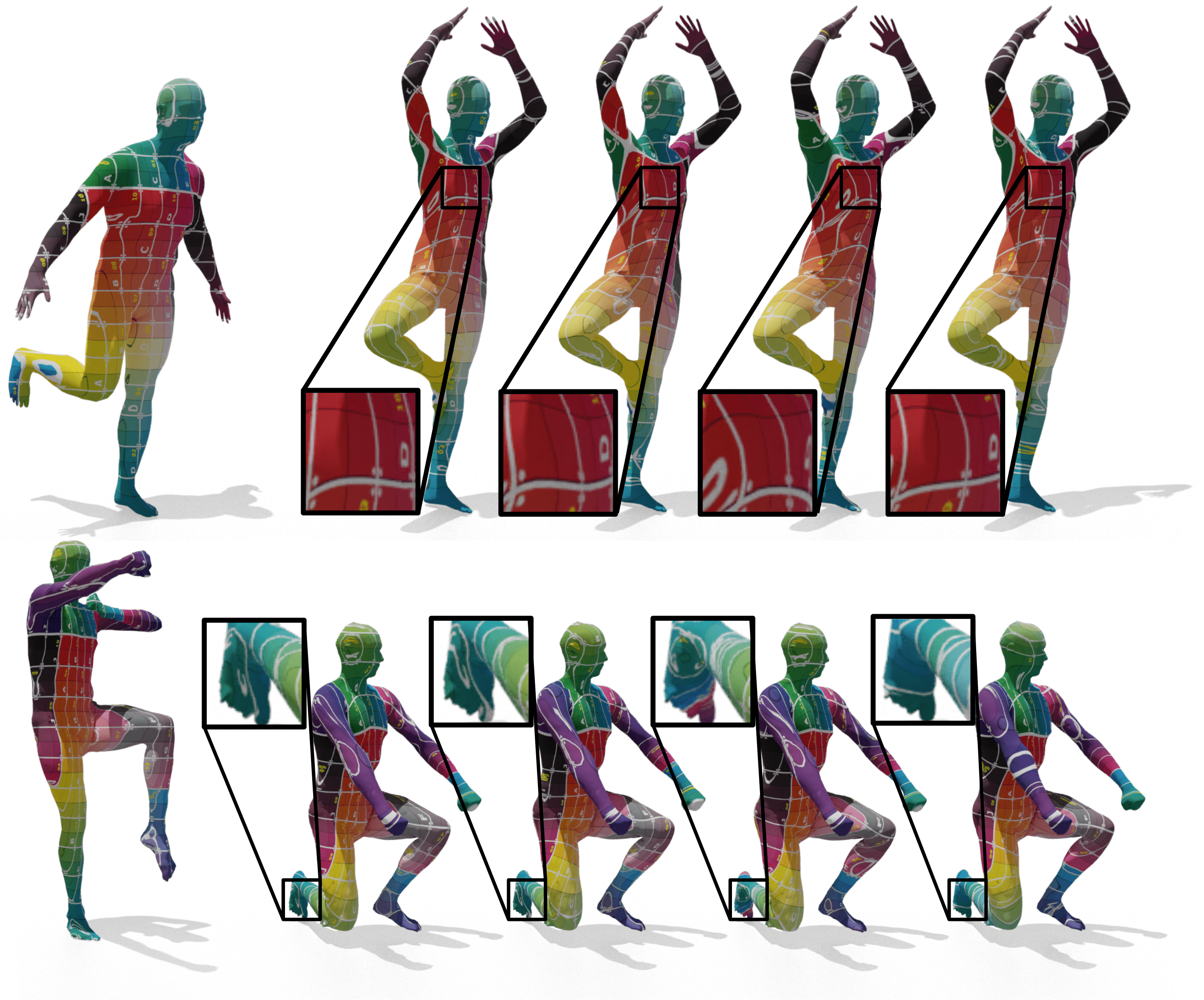}
		
		\put(5,0){\scriptsize{$\M$}}
		\put(30,0){\scriptsize{GT}}
		\put(45,0){\scriptsize{Our}}
		\put(56,0){\scriptsize{$\DiffNet$}}
		\put(80,0){\scriptsize{Universal}}
		
		\put(90,65){\scriptsize{S+F}}
		\put(90,25){\scriptsize{F+S}}
		\end{overpic}
		
		\caption{\label{fig:matching} Texture transfer for matching quality comparison.}
\end{figure*}

In Figure \ref{fig:matching}, we show texture transfer on S+F and F+S. While all methods produce an excellent matching, we observe a misalignment of high-frequency details. 

\begin{figure}
	\centering
	\footnotesize
	
	\begin{overpic}[trim=0cm 0cm 0cm 0cm,clip, width=0.93\linewidth]{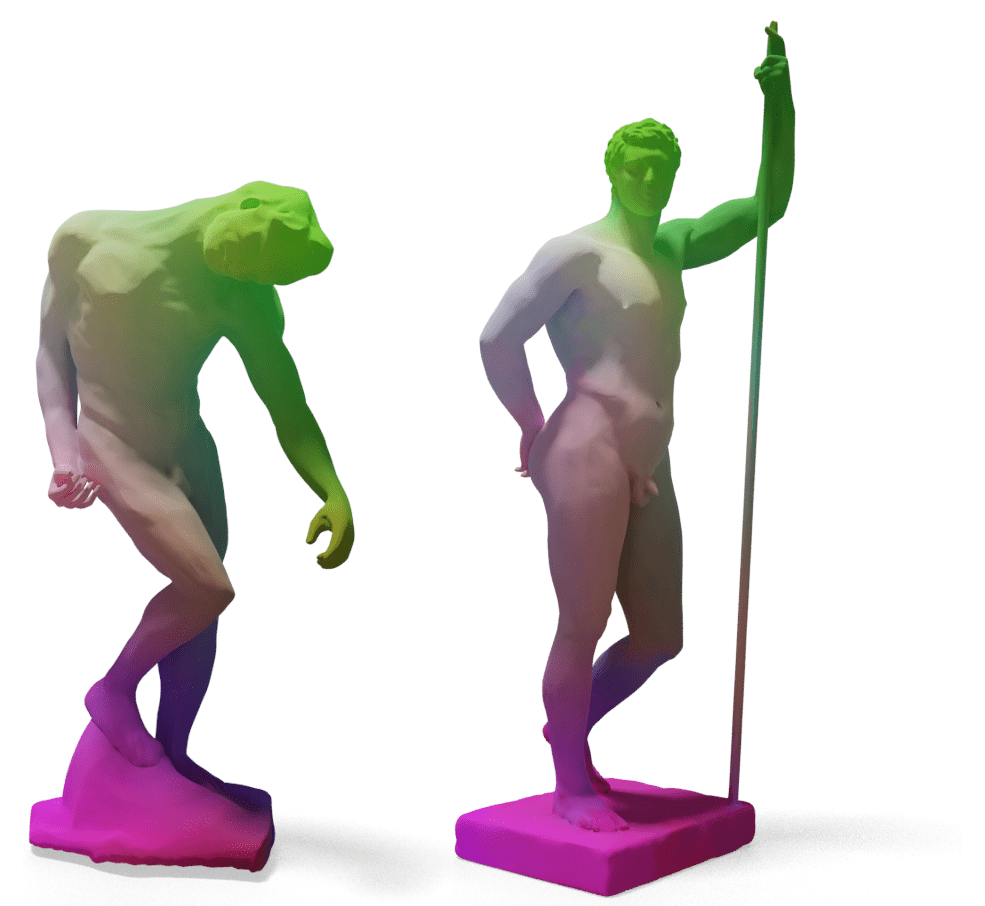}

	\end{overpic}
	\vspace{-0.3cm}
	
	\caption{\label{fig:statues}  An example of extreme non-isometric matching. The two shapes are from \cite{scantheworld}.}
\hspace{-0.2cm}

\end{figure}

\subsection{Qualitative results}
\begin{figure*}[!t]
	\centering
	\footnotesize
	
	\begin{overpic}[trim=0cm 0cm 0cm 0cm,clip, width=0.93\linewidth]{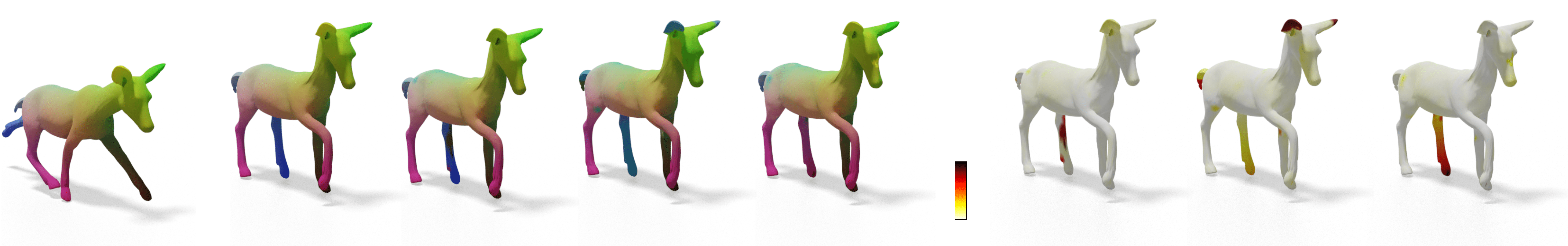}
		\put(5,16){$\M$}
		
		\put(62,1){\scriptsize{0}}
		\put(62,4){\scriptsize{max}}

		\put(15,16){\scriptsize{GT}}
		\put(25,16){\scriptsize{Our$^\EmbNet$}}
		\put(35,16){\scriptsize{$\DiffNet$}}
		\put(50,16){\scriptsize{Uni}}
		
		\put(65,16){\scriptsize{Our$^\EmbNet$}}
		\put(78,16){\scriptsize{$\DiffNet$}}
		\put(90,16){\scriptsize{Uni}}
	\end{overpic}
	\vspace{-0.3cm}
	
	\caption{\label{fig:smal2} An isometric pair from the remeshed SMAL dataset. From the left: the source shape, the target shape with color transferred using $\Pi_{\M\N}^{gt}$. Then, the color transfer is performed by our, $\DiffNet$, and the Universal Embedding Baseline. On the right, the geodesic error depicted over the surface. }
\hspace{-0.2cm}

\end{figure*}

\begin{figure*}[!t]
	\centering
	\footnotesize
	
	\begin{overpic}[trim=0cm 0cm 0cm 0cm,clip, width=0.93\linewidth]{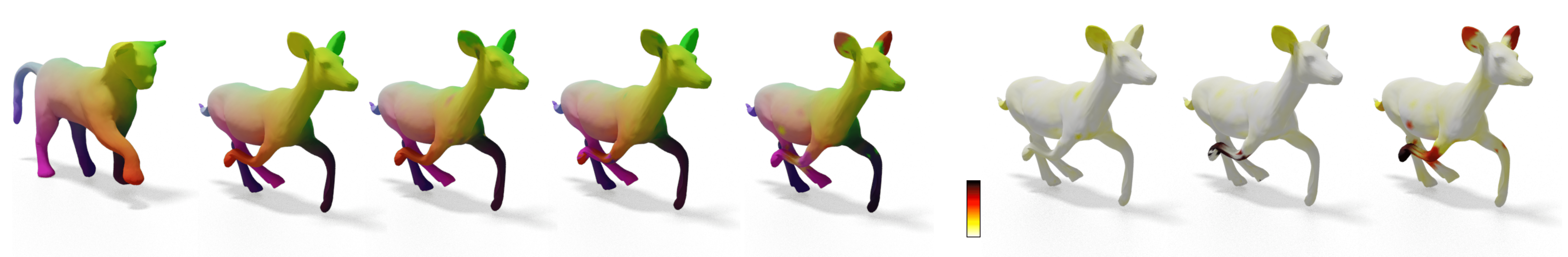}
		\put(5,16){$\M$}
		
		\put(63,1){\scriptsize{0}}
		\put(63,4){\scriptsize{max}}

		\put(15,16){\scriptsize{GT}}
		\put(25,16){\scriptsize{Our$^\EmbNet$}}
		\put(35,16){\scriptsize{$\DiffNet$}}
		\put(50,16){\scriptsize{Uni}}
		
		\put(65,16){\scriptsize{Our$^\EmbNet$}}
		\put(78,16){\scriptsize{$\DiffNet$}}
		\put(90,16){\scriptsize{Uni}}
	\end{overpic}
	
	\caption{\label{fig:smal} A non-isometric pair from the remeshed SMAL dataset. From the left: the source shape, the target shape with color transferred using $\Pi_{\M\N}^{gt}$. Then, the color transfer performed by our, $\DiffNet$, and the Universal Baseline. On the right, the geodesic error depicted over the surface. }
\hspace{-0.2cm}

\end{figure*}

\begin{figure*}
	\centering
	\footnotesize
	
	\begin{overpic}[trim=0cm 0cm 0cm 0cm,clip, width=0.93\linewidth]{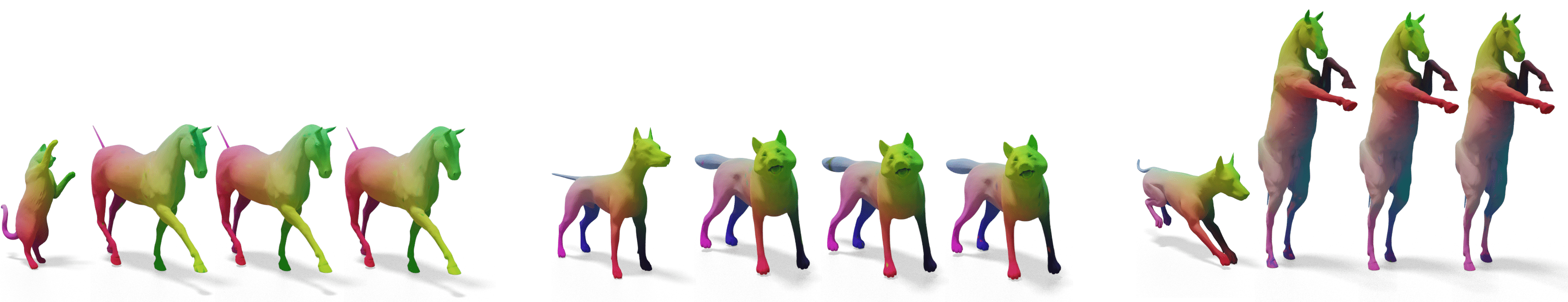}
		\put(5,0){$\M$}
		\put(15,0){Our$^\EmbNet$}
		
		\put(38,0){$\M$}
		\put(60,0){Our$^\EmbNet$}
		
		\put(80,0){$\M$}
		\put(93,0){Our$^\EmbNet$}		
	\end{overpic}
	
	\caption{\label{fig:toscasup} Three pair of animals from TOSCA \cite{TOSCA} dataset, showing our generalization capability.}
\hspace{-0.2cm}

\end{figure*}

\begin{figure*}

		\begin{overpic}[trim=0cm 0cm 0cm 0cm,clip, width=0.95\linewidth]{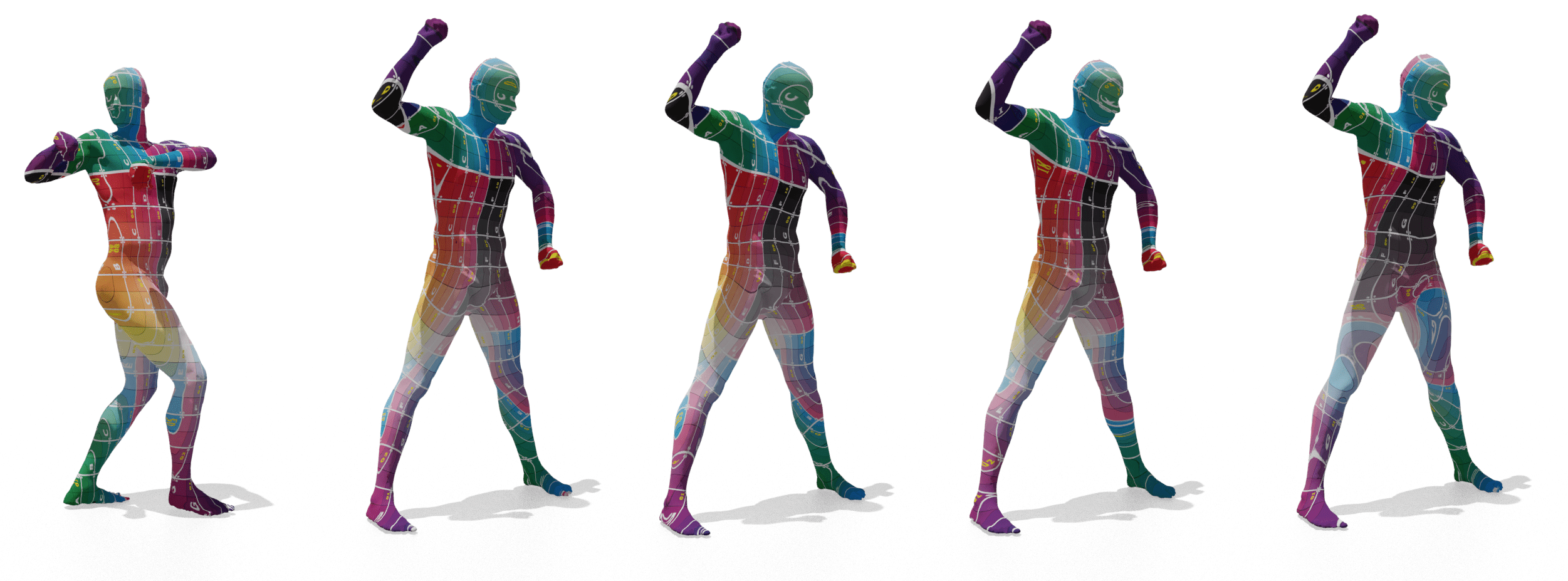}
		
		\put(5,0){\scriptsize{$\M$}}
		\put(30,0){\scriptsize{GT}}
		\put(45,0){\scriptsize{Our$^\EmbNet$}}
		\put(63,0){\scriptsize{$\DiffNet$}}
		\put(88,0){\scriptsize{Universal}}
		
		\end{overpic}
		
		\caption{\label{fig:matching1} Texture transfer for matching quality comparison (F+S setting).}
\end{figure*}

\begin{figure*}

		\begin{overpic}[trim=0cm 0cm 0cm 0cm,clip, width=0.95\linewidth]{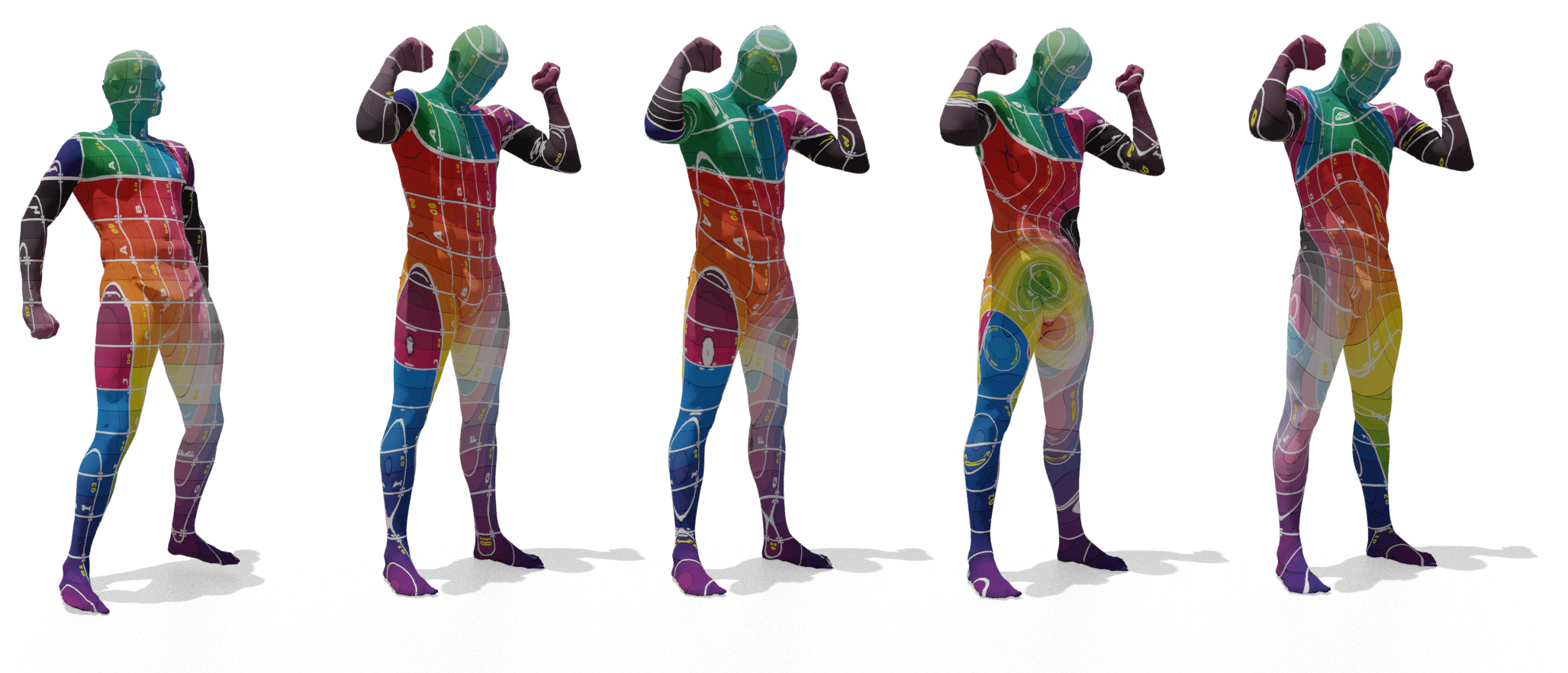}
		
		\put(5,0){\scriptsize{$\M$}}
		\put(30,0){\scriptsize{GT}}
		\put(45,0){\scriptsize{Our$^\EmbNet$}}
		\put(63,0){\scriptsize{$\DiffNet$}}
		\put(88,0){\scriptsize{Universal}}
		
		\end{overpic}
		\vspace{-0.2cm}
		
		\caption{\label{fig:matching2} Texture transfer for matching quality comparison (F+S setting).}
\end{figure*}

\begin{figure*}

		\begin{overpic}[trim=0cm 0cm 0cm 0cm,clip, width=0.95\linewidth]{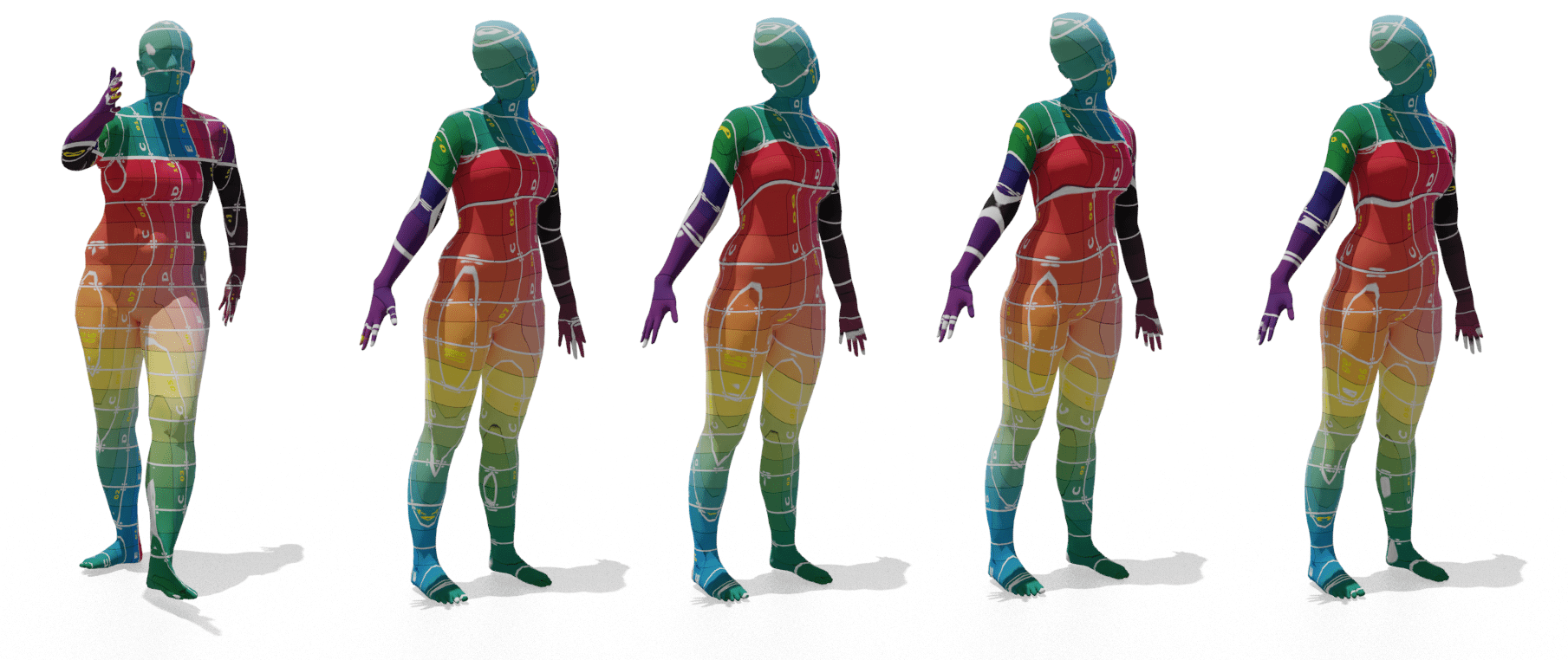}
		
		\put(5,0){\scriptsize{$\M$}}
		\put(30,0){\scriptsize{GT}}
		\put(45,0){\scriptsize{Our$^\EmbNet$}}
		\put(63,0){\scriptsize{$\DiffNet$}}
		\put(88,0){\scriptsize{Universal}}
		
		\end{overpic}
		
		\caption{\label{fig:matching3} Texture transfer for matching quality comparison (S+F setting).}
\end{figure*}

\begin{figure*}

		\begin{overpic}[trim=0cm 0cm 0cm 0cm,clip, width=0.95\linewidth]{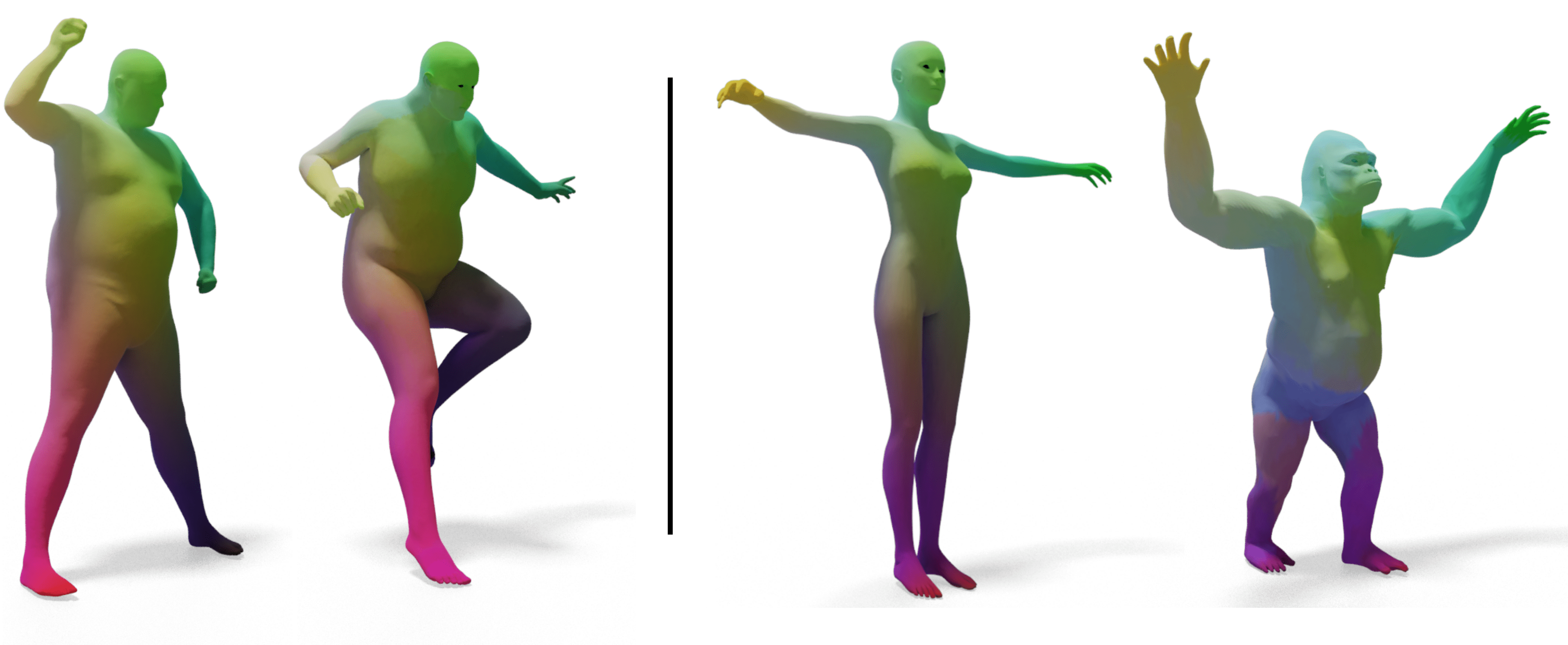}
		\put(5,0){\scriptsize{$\M$}}
		\put(25,0){\scriptsize{Our$^\EmbNet$}}
		
		\put(57,0){\scriptsize{$\M$}}
		\put(80,0){\scriptsize{Our$^\EmbNet$}}	
		\end{overpic}
		
		\caption{\label{fig:outdis} Color transfer for matching quality comparison on two out of distributions couples.}
\end{figure*}
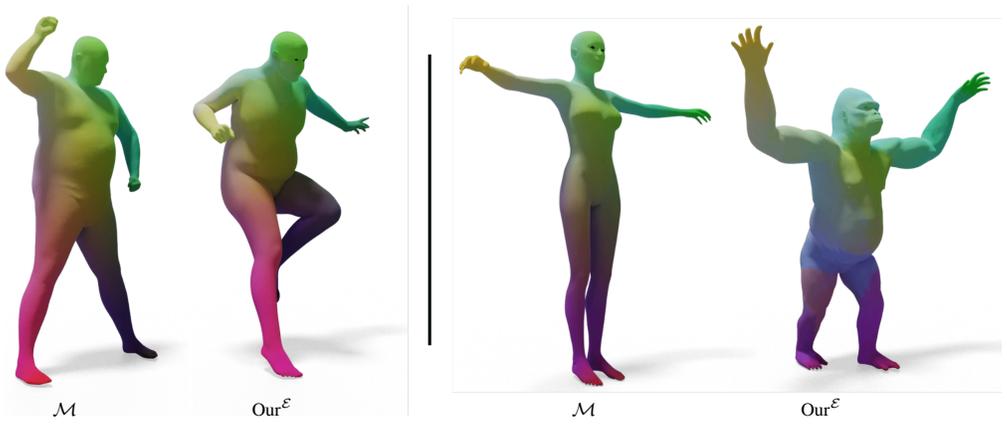

Here we report some additional qualitative results:
\begin{enumerate}
    \item In Figure \ref{fig:statues} we report another example of our matching on two statues. Notice that they do not share the same topology structure: despite this, we obtain a smooth matching.
    \item In Figure \ref{fig:smal2} we report an example between two similar animals. Even in the isometric case (which is favorable to the LBO) we show better results.
    \item In Figure \ref{fig:smal} we have a significant non-isometric case; the error is localized on protrusions (ears, legs).
    \item In Figure \ref{fig:toscasup} we show our results on three pairs of animals from the TOSCA dataset \cite{TOSCA}. We matched highly non-isometric shapes to test our generalization capability. Notice that cats and horses are classes not seen at training time.
    \item in Figures \ref{fig:matching1}, \ref{fig:matching2}, and \ref{fig:matching3} we show three more qualitative results. We would like to highlight that our methods seems to obtain better matching on protrusions (i.e., legs and arms).
    \item in Figure \ref{fig:outdis} we show an example on two further different datasets. On the left, a pair from SHREC19 \cite{SHREC19}; notice that the two shapes are different for their pose but also for their quality (i.e., the $\M$ is a real scan, while the target one is a synthetic model). On the right, a woman is matched with a gorilla from TOSCA dataset \cite{TOSCA}. We consider this case particularly extreme since arms and legs have entirely different proportions. Even if there are some evident artifacts on the stomach, we observe an overall coherence in the obtained matching.
\end{enumerate}

\section{ZoomOut}
\label{sec:ZoomOut}

\subsection{Can we refine the learned embeddings?}

\begin{table*}[!t]
\centering
\footnotesize
	\ra{1.2}
	\begin{tabular}{@{}rccccccccccc@{}}
		& \multicolumn{5}{c}{S+F}  & \phantom{abc} & \multicolumn{5}{c}{F+S}  \\
		\cmidrule{2-6}  \cmidrule{8-12} 
		&             Init         &   ZO10       & ZO20             & ZO40 & ZO60           &    & Init                &  ZO10           & ZO20          & ZO40            & ZO60   \\ \midrule

		$\mathbf{I_{Our}}$ $\OurBasis$ & \multirow{2}{*}{ 3.77}  & \textbf{3.18}  & \textbf{3.16}       & \textbf{3.18}       & \textbf{3.21}         &&\multirow{2}{*}{8.76}     & \textbf{7.33}       & \textbf{7.20}     & \textbf{7.30}  & \textbf{7.30}   \\
		$\mathbf{I_{Our}}$ $\LBOBasis$    &                          &  3.98            & 4.02 & 4.13       & 4.17                   &                  &       &    8.32             &  8.58             &   8.58 & 8.71  \\ \hline
		$\mathbf{I_{\DiffNet}} $ $\OurBasis$    &  \multirow{2}{*}{ 2.85}  &  \textbf{2.55} & \textbf{2.54}  & \textbf{2.57} & \textbf{2.62}       & & \multirow{2}{*}{10.17}  & \textbf{8.99}  &  \textbf{8.97} & \textbf{9.03}          & \textbf{9.11} \\
		$\mathbf{I_{\DiffNet}} $ $\LBOBasis$       &                          &  2.96          & 3.04           &  3.12     & 3.18                     &    &                       &  9.63         & 9.72          &  10.03 & 10.08         \\
		\bottomrule
	\end{tabular}
	\caption{Matching results using ZoomOut as further refinement of an input matching. Replacing the first $40$ $\LBOBasis$ with $\OurBasis$, we obtain a better matching even for the initialization provided by $\DiffNet$.  }
	\label{tab:ZO}
\end{table*}

\begin{figure*}[!t]
\centering
\begin{overpic}[width=0.92\textwidth]{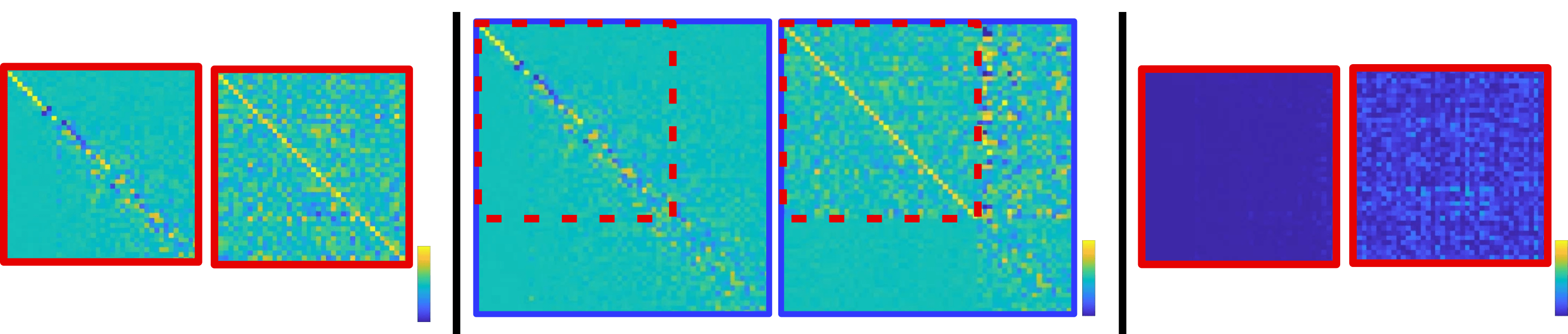}
\put(26,-0.5){\tiny{$-1$}}
\put(27,6){\tiny{$1$}}

\put(69,-0.5){\tiny{$-1$}}
\put(70,6){\tiny{$1$}}

\put(100,-.05){\tiny{$0$}}
\put(100,6){\tiny{$1$}}

\put(13,-0.5){\scriptsize{$C_{init}$}}
\put(4,18){\scriptsize{$\LBOBasis$}}
\put(20,18){\scriptsize{$\OurBasis$}}

\put(48,-0.5){\scriptsize{ $C_{ZO}$}}
\put(40,21){\scriptsize{$\LBOBasis$}}
\put(60,21){\scriptsize{$\OurBasis$}}

\put(77,-0.5){\scriptsize{$\|C_{init} - C_{ZO_{20\times 20}}  \|$}}
\put(78,18){\scriptsize{$\LBOBasis$}}
\put(93,18){\scriptsize{$\OurBasis$}}

\end{overpic}
\caption{\label{fig:cdiff} ZoomOut applied to $\OurBasis$ and $\LBOBasis$. On the left, the two transformation matrices have been initialized by the same correspondence $\Pi_{\M\N}$. Then, in the middle, we increased their size by $20$ dimensions, using in both cases the LBO basis from $41$ to $60$. We show the absolute difference between the initialization and the new upper right part of the matrix on the right. }
	\hspace{-3.5cm}
	
\end{figure*}

One of the main advantages of a spectral embedding is its arbitrary dimensionality, which can be selected at test time. 
However, in \cite{LIE2020} the basis dimension is fixed a priori.
Inspired by this, we propose to incorporate ZoomOut~\cite{melzi2019zoomout} in our pipeline, considering our $40$-dimensional embedding equivalent to the first $40$ eigenfunctions of the LBO. Then, we apply ZoomOut by introducing LBO eigenfunctions (starting from the $41$st). 
In the first two rows of Table \ref{tab:ZO}, we initialize with the matching produced by our method the $C \in \mathbb{R}^{40 \times 40}$  ($I_{Our})$ exploiting our basis ($1$st row) or the first $40$ LBO basis functions ($2$nd row).
Then, we applied ZoomOut to increase the set with higher frequencies (adding 5 basis functions at each iteration in the ZoomOut pipeline). 
We observe that our basis can replace the low frequencies of the LBO.
In the $3$rd and $4$th rows, we repeated the experiment initializing with the matching provided by $\DiffNet$ ($I_{\DiffNet}$), improving a matching optimized for another basis.
ZoomOut shows instability for some dimensionality while, remarkably, our embedding seems more stable.

We depict in Figure \ref{fig:cdiff} the different effects of ZoomOut on the linear transformation. Starting with $\Pi_{\M\N}$, we obtain the $40 \times 40$ transformation using either our embedding ($ C^{\OurBasis}_{init}$) or the first $40$ LBO basis functions ($ C^{\LBOBasis}_{init}$). Then, we applied ZoomOut obtaining $C^{\OurBasis}_{ZO}$ and $C^{\LBOBasis}_{ZO}$ of dimension $60 \times 60$. We observe that $C^{\OurBasis}_{ZO}$ presents a block structure: the top-left block highlights how $\OurBasis$ have been recombined to match higher frequencies of $\LBOBasis$; the bottom-left block is almost $0$. 
On the right, we report the variations in the upper left $20 \times 20$; the $\LBOBasis$ one is left unchanged, while our basis shows flexibility thanks to its non-orthogonality.

\begin{figure*}[!t]
\begin{overpic}[width=0.98\textwidth]{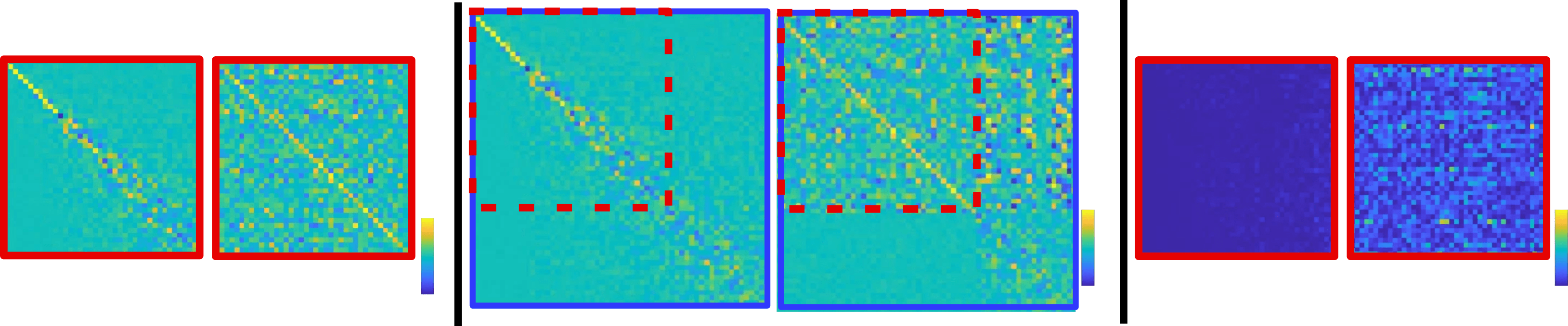}
\put(26,0.5){\tiny{$-1$}}
\put(27,7){\tiny{$1$}}

\put(69,0.5){\tiny{$-1$}}
\put(70,7){\tiny{$1$}}

\put(100,0.5){\tiny{$0$}}
\put(100,6){\tiny{$0.5$}}

\put(13,0.5){\scriptsize{$C_{init}$}}
\put(6,21){\scriptsize{$\LBOBasis$}}
\put(19,21){\scriptsize{$\OurBasis$}}

\put(48,-0.5){\scriptsize{ $C_{ZO}$}}
\put(40,21){\scriptsize{$\LBOBasis$}}
\put(60,21){\scriptsize{$\OurBasis$}}

\put(77,-0.5){\scriptsize{$\|C_{init} - C_{ZO_{20\times 20}}  \|$}}
\put(78,21){\scriptsize{$\LBOBasis$}}
\put(93,21){\scriptsize{$\OurBasis$}}

\end{overpic}
\caption{\label{fig:cdiff1} ZoomOut applied to $\OurBasis$ and $\LBOBasis$. On the left, the two transformation matrices have been initialized by the same correspondence $\Pi_{\M\N}$. Then, in the middle, we increased their size of $20$ dimensions, using in both cases the LBO basis from $41$ to $60$. On the right, we show the absolute difference between the initialization and the new upper left part of the matrix. The considered example refers to the pair shown in Figure \ref{fig:matching1}.}
	\hspace{-3.5cm}
	
\end{figure*}

\begin{figure*}[!t]
\begin{overpic}[width=0.98\textwidth]{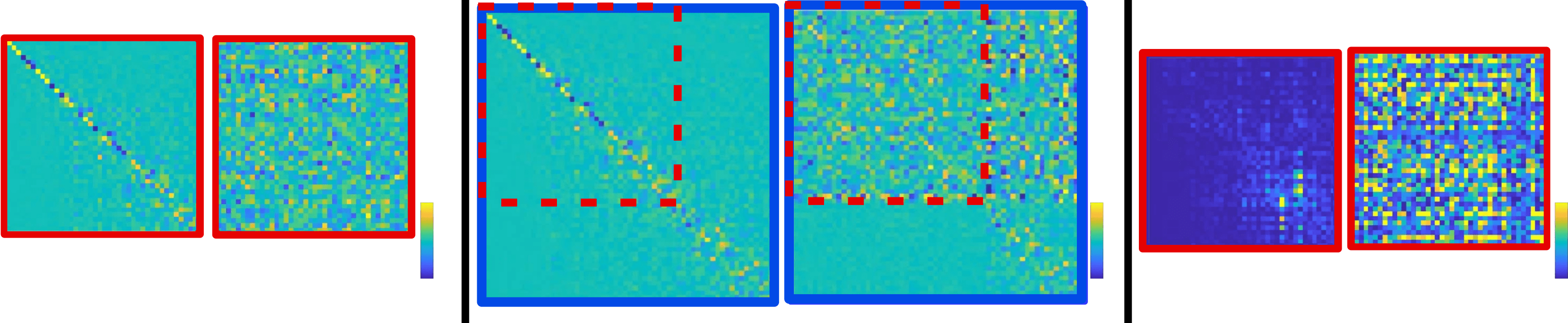}
\put(26,1){\tiny{$-1$}}
\put(27,8){\tiny{$1$}}

\put(69,1){\tiny{$-1$}}
\put(70,8){\tiny{$1$}}

\put(100,1){\tiny{$0$}}
\put(100,8){\tiny{$0.5$}}

\put(13,-0.5){\scriptsize{$C_{init}$}}
\put(4,21){\scriptsize{$\LBOBasis$}}
\put(20,21){\scriptsize{$\OurBasis$}}

\put(48,-0.5){\scriptsize{ $C_{ZO}$}}
\put(40,21){\scriptsize{$\LBOBasis$}}
\put(60,21){\scriptsize{$\OurBasis$}}

\put(77,-0.5){\scriptsize{$\|C_{init} - C_{ZO_{20\times 20}}  \|$}}
\put(78,21){\scriptsize{$\LBOBasis$}}
\put(93,21){\scriptsize{$\OurBasis$}}

\end{overpic}
\caption{\label{fig:cdiff2} ZoomOut applied to $\OurBasis$ and $\LBOBasis$. On the left, the two transformation matrices have been initialized by the same correspondence $\Pi_{\M\N}$. Then, in the middle, we increased their size of $20$ dimensions, using in both cases the LBO basis from $41$ to $60$. On the right, we show the absolute difference between the initialization and the new upper left part of the matrix. The considered example refers to the pair shown in Figure \ref{fig:matching2}.}
	\hspace{-3.5cm}
	
\end{figure*}

In Figures \ref{fig:cdiff1} and \ref{fig:cdiff2} we report two other cases of ZoomOut and their impact on the initialization matrix (the considered pairs are the ones of Figure \ref{fig:matching1} and \ref{fig:matching2} respectively). In the first one, we observe that the initial LBO matrix has a less diagonal behaviour than the latter. This is in general due to a non-isometric deformation that in this case, could be reasonably caused by a twist of the torso of Figure \ref{fig:matching1}. The $C_{init}$ seems to be almost preserved by the ZoomOut process. In the second case, the initialization is more diagonal, witnessing two almost isometrical shapes. In this case, the ZoomOut process can recombine a few parts of the LBO $C_{init}$, while this impacts mainly the last dimensions. However, in both cases, our matrix shows better flexibility. Also, in all experiments, we observe the block division of our $C_{ZO}$ matrix discussed above.

\paragraph{Insights.} This experiment shows that learned embeddings are a good initialization for refinement techniques and improve existing learning pipelines. The structure of the final $C$ matrix also reveals the reason behind this. In the obtained matrix using only $\LBOBasis$, the top right and bottom left rectangles are almost empty, and the only significant interaction appears in the added frequencies. Considering the $C$ produced with $\OurBasis$, this interaction is almost unchanged. However, the top right block shows the interaction between $\OurBasis$ and higher frequencies. This opens to a new perspective in the field of shape matching: instead of seeking competition between representations, incorporating the two seems a promising direction. Note that our merging is naive, and more sophisticated combinations are possible; we leave this analysis for future work.

\subsection{Step size ablation study}
In Table \ref{tab:ZO}, we reported the results by using ZoomOut and step size $5$ (i.e., including $5$ more basis at each ZoomOut iteration). Here, we report also results using a step of $1$ (Table \ref{tab:ZO1}) and $2$ (Table \ref{tab:ZO2}). We show that our basis is always a better choice to improve the matching. We noticed that by reducing the step size, the results at higher $C$ dimensions become more unstable. We believe this is due to the more iterations required to reach the same $C$ dimensions, introducing noise in the correspondence.

\begin{table*}\centering
\scriptsize
	\ra{1.2}
	\begin{tabular}{@{}rccccccccccc@{}}\toprule
		& \multicolumn{5}{c}{S+F}  & \phantom{abc} & \multicolumn{5}{c}{F+S}  \\
		\cmidrule{2-6}  \cmidrule{8-12} 
		&             Init         &   ZO10       & ZO20             & ZO40 & ZO60           &    & Init                &  ZO10           & ZO20          & ZO40            & ZO60   \\ \midrule

		$\mathbf{I_{Our}}$ $\OurBasis$ & \multirow{2}{*}{ 3.77}  & \textbf{3.45}  & \textbf{3.45}       & \textbf{3.45}       & \textbf{3.67}         &&\multirow{2}{*}{8.76}     & \textbf{7.78}       & \textbf{7.50}     & \textbf{7.37}  & \textbf{7.42}   \\
		$\mathbf{I_{Our}}$ $\LBOBasis$    &                          &  3.81            & 4.11 & 4.55       & 4.82                   &                  &       &    8.44             &  8.66             &   8.86 & 8.99  \\ \hline
		$\mathbf{I_{\DiffNet}} $ $\OurBasis$    &  \multirow{2}{*}{ 2.85}  &  \textbf{2.88} & \textbf{2.93}  & \textbf{3.00} & \textbf{3.22}       & & \multirow{2}{*}{10.17}  & \textbf{9.44}  &  \textbf{9.35} & \textbf{9.43}          & \textbf{9.38} \\
		$\mathbf{I_{\DiffNet}} $ $\LBOBasis$       &                          &  2.92          & 3.28           &  3.70     & 3.95                     &    &                       &  9.72         & 9.89          &  10.27 & 10.41         \\
		\bottomrule
	\end{tabular}
	\caption{Matching results using ZoomOut with step size $1$.}
	\label{tab:ZO1}
\end{table*}

\begin{table*}\centering
\scriptsize
	\ra{1.2}
	\begin{tabular}{@{}rccccccccccc@{}}\toprule
		& \multicolumn{5}{c}{S+F}  & \phantom{abc} & \multicolumn{5}{c}{F+S}  \\
		\cmidrule{2-6}  \cmidrule{8-12} 
		&             Init         &   ZO10       & ZO20             & ZO40 & ZO60           &    & Init                &  ZO10           & ZO20          & ZO40            & ZO60   \\ \midrule

		$\mathbf{I_{Our}}$ $\OurBasis$ & \multirow{2}{*}{ 3.77}  & \textbf{3.34}  & \textbf{3.36}       & \textbf{3.52}       & \textbf{3.54}         &&\multirow{2}{*}{8.76}     & \textbf{7.54}       & \textbf{7.35}     & \textbf{7.27}  & \textbf{7.28}   \\
		$\mathbf{I_{Our}}$ $\LBOBasis$    &                          &  3.89            & 4.21 & 4.49       & 4.49                   &                  &       &    8.37             &  8.61             &   8.75 & 8.94  \\ \hline
		
		$\mathbf{I_{\DiffNet}} $ $\OurBasis$    &  \multirow{2}{*}{ 2.85}  &  \textbf{2.76} & \textbf{2.83}  & \textbf{3.03} & \textbf{3.03}       & & \multirow{2}{*}{10.17}  & \textbf{9.28}  &  \textbf{9.18} & \textbf{9.22}          & \textbf{9.34} \\
		$\mathbf{I_{\DiffNet}} $ $\LBOBasis$       &                          &  2.97          & 3.25           &  3.51     & 3.56                     &    &                       &  9.65         & 9.99          &  10.19 & 10.48         \\
		\bottomrule
	\end{tabular}
	\caption{Matching results using ZoomOut with step size $2$. }
	\label{tab:ZO2}
\end{table*}

\end{document}